\documentclass{article}
\usepackage{arxiv}

\usepackage[numbers]{natbib}

\usepackage[document,symbol]{ragged2e}
\usepackage{amsfonts}
\usepackage{amssymb}
\usepackage{amsmath}
\usepackage{amsthm}
\usepackage{pict2e}
\usepackage{bm}
\usepackage{siunitx}
\usepackage{caption}
\usepackage{algorithm}
\usepackage[noend]{algpseudocode}
\usepackage{xcolor}
\usepackage{array}
\usepackage{url}
\usepackage{pifont}
\usepackage{graphicx}
\usepackage{hyperref}
\usepackage{booktabs}
\usepackage{colortbl}
\usepackage{multirow}
\usepackage[symbol]{footmisc}
\usepackage{caption}
\usepackage{comment}
\usepackage{fancyhdr}
\usepackage{afterpage}

\newcommand{\R}{\mathbb{R}}
\newcommand{\N}{\mathbb{N}}
\DeclareMathOperator*{\argmin}{arg\,min} 

\newcommand{\norm}[1]{\lVert#1\rVert}
\newcommand{\Norm}[1]{\left\lVert#1\right\rVert}
\newcommand\mytilde[1]{\stackrel{\sim}{\smash{#1}\rule{0pt}{1.05ex}}}

\newcolumntype{L}[1]{>{\raggedright\let\newline\\\arraybackslash\hspace{0pt}}m{#1}}
\newcolumntype{C}[1]{>{\centering\let\newline\\\arraybackslash\hspace{0pt}}m{#1}}
\newcolumntype{R}[1]{>{\raggedleft\let\newline\\\arraybackslash\hspace{0pt}}m{#1}}

\newcommand\Tstrut{\rule{0pt}{2.6ex}}

\newtheorem{remark}{Remark}
\newtheorem*{remark*}{Remark}

\graphicspath{ {img/} }
	
\title{Deformable registration and generative\\ modelling of aortic anatomies by\\ auto--decoders and neural ODEs}

\author{}

\date{}

\renewcommand{\headeright}{Deformable registration and generative modelling of aortic anatomies by ADs and NODEs}
\renewcommand{\undertitle}{}

\afterpage{\cfoot{\thepage}}

\begin{document}
	
\maketitle

\vspace{-2.5cm}

\begin{center}
\textbf{Riccardo Tenderini\textsuperscript{1,}\footnote{\emph{Corresponding author.} Email: \href{mailto:riccardo.tenderini@outlook.com}{\texttt{riccardo.tenderini@outlook.com}}} \hspace{.5 cm} Luca Pegolotti\textsuperscript{2,3,4} \hspace{.5cm} Fanwei Kong\textsuperscript{3,4,5} \hspace{.5 cm} Stefano Pagani\textsuperscript{6}\\[10pt]
Francesco Regazzoni\textsuperscript{6} \hspace{.5cm} Alison L. Marsden\textsuperscript{2,3,4,7} \hspace{.5 cm} Simone Deparis\textsuperscript{1}}

\bigskip

\textsuperscript{1}{Institute of Mathematics, EPFL, Lausanne, Switzerland}

\textsuperscript{2}{Department of Bioengineering, Stanford University, CA, USA}

\textsuperscript{3}{Department of Pediatrics, Stanford University, CA, USA}

\textsuperscript{4}{Institute for Computational and Mathematical Engineering, Stanford University, CA, USA}

\textsuperscript{5}{Department of Mechanical Engineering and Materials Science, Washington University, St. Louis, MO, USA}

\textsuperscript{6}{MOX -- Department of Mathematics, Politecnico di Milano, Milano, Italy}

\textsuperscript{7}{Cardiovascular Institute, Stanford University, CA, USA}

\end{center}

\bigskip\bigskip

\begin{abstract}
\justifying
This work introduces AD--SVFD, a deep learning model for the deformable registration of vascular shapes to a pre--defined reference and for the generation of synthetic anatomies. AD--SVFD operates by representing each geometry as a weighted point cloud and models ambient space deformations as solutions at unit time of ODEs, whose time--independent right--hand sides are expressed through artificial neural networks. The model parameters are optimized by minimizing the Chamfer Distance between the deformed and reference point clouds, while backward integration of the ODE defines the inverse transformation.
A distinctive feature of AD--SVFD is its auto--decoder structure, that enables generalization across shape cohorts and favors efficient weight sharing. In particular, each anatomy is associated with a low--dimensional code that acts as a self--conditioning field and that is jointly optimized with the network parameters during training. At inference, only the latent codes are fine--tuned, substantially reducing computational overheads.
Furthermore, the use of implicit shape representations enables generative applications: new anatomies can be synthesized by suitably sampling from the latent space and applying the corresponding inverse transformations to the reference geometry. Numerical experiments, conducted on healthy aortic anatomies, showcase the high--quality results of AD--SVFD, which yields extremely accurate approximations at competitive computational costs.
\end{abstract}


\maketitle

\justifying

\section*{Introduction}\label{sec:introduction}

Over the last two decades, the deformable registration of three--dimensional images has become increasingly important in a wide number of computer graphics and computer vision applications. In broad terms, the deformable --- or non--rigid --- registration problem consists in aligning and locating different shapes within a shared coordinate system, to enable meaningful comparisons and analyses~\cite{modersitzki2003numerical, modersitzki2009fair, ashburner2007fast}. 
Besides industrial and engineering applications, deformable registration nowadays plays a crucial role in several medical imaging tasks, such as multimodal image fusion, organ atlas creation, and monitoring of disease progression~\cite{sotiras2013deformable,haskins2020deep}.
Unlike rigid registration, which involves only global scaling, rotations, and translations, deformable registration must estimate complex, localized deformation fields that account for natural anatomical variability. This challenge is enhanced by the presence of noise, outliers, and partial overlaps, which are very common in clinical data. Furthermore, exact point--to--point correspondences between different anatomies are rarely available in practice, which requires the adoption of alternative metrics to evaluate data adherence. \\

The challenge of developing efficient, reliable, and computationally tractable registration methods is of paramount importance for improving medical imaging workflows, healthcare technologies, and patient care. 
Manual alignment of images in subject--specific clinical contexts is often infeasible or impractical, due to the complexity and variability of biological structures, as well as to the differences in imaging modalities and acquisition times. To address this limitation, several automatic registration approaches have been developed. Among the most widely employed ones, we can mention DARTEL~\cite{ashburner2007fast}, Diffeomorphic Demons~\cite{vercauteren2009diffeomorphic}, and LDDMM~\cite{beg2005computing, vaillant2005surface, bone2018deformetrica}. Notably, all these methods share remarkable robustness characteristics, since they are based on a deformation of the ambient space, which is guaranteed to be smooth, differentiable, invertible, and topology preserving. \\

While traditional image and shape registration approaches can yield extremely accurate results, they nonetheless entail non--negligible computational costs, that may hinder their use in real--time clinical practice. To mitigate this issue and improve the overall performance, deep learning (DL) techniques have been exploited in various ways. A non--exhaustive list of the most popular state--of--the--art DL--based registration methods includes the probabilistic models developed in \cite{krebs2018unsupervised, dalca2019unsupervised}, \emph{Voxelmorph}~\cite{balakrishnan2019voxelmorph}, \emph{Smooth Shells}~\cite{eisenberger2020smooth}, \emph{Neuromorph}~\cite{eisenberger2021neuromorph}, \emph{Cyclemorph}~\cite{kim2021cyclemorph}, \emph{Diffusemorph}~\cite{kim2022diffusemorph} and \emph{Transmorph}~\cite{chen2022transmorph}.
We refer to~\cite{haskins2020deep, zou2022review, deng2022survey, ramadan2024medical} for comprehensive literature reviews on the topic. \\

In our study, we are specifically interested in the registration of vascular surfaces. The latter can be seamlessly extracted from volumetric data, acquired through traditional imaging modalities, such as CT--scans or MRI. Furthermore, novel techniques like photoacoustic scanning~\cite{laufer2012vivo, huang2023dual, huynh2024fast} are rapidly gaining traction in clinical practice, since they provide a low--cost radiation--free alternative, particularly well--suited for superficial vascular anatomies, located up to $15 \ \si{\milli\metre}$ beneath the skin.
A review of the classical techniques for surface registration can be found in~\cite{tam2012registration}. In this scenario, DL--based approaches can be subdivided into two major groups, depending on how surfaces are represented. On the one hand, we have methods that treat shapes as 3D point clouds~\cite{monji2023review}, such as the ones introduced in \cite{liu2019flownet3d, croquet2021unsupervised, amor2022resnet}. On the other hand, instead, there exist several methods that represent 3D geometries by means of Deep Implicit Functions --- namely continuous signed distance functions, expressed through neural networks~\cite{chen2019learning, xu2019disn, park2019deepsdf} --- such as the ones presented in \cite{sun2022topology, kong2024sdf4chd}. Notably, the models described in \cite{croquet2021unsupervised, sun2022topology, kong2024sdf4chd} encapsulate learnable latent shape representations, which enable the simultaneous registration of multiple geometries to a common reference, as well as their use as generative AI tools. \\  

\begin{figure}[t!]
\centering
\includegraphics[width=.99\textwidth]{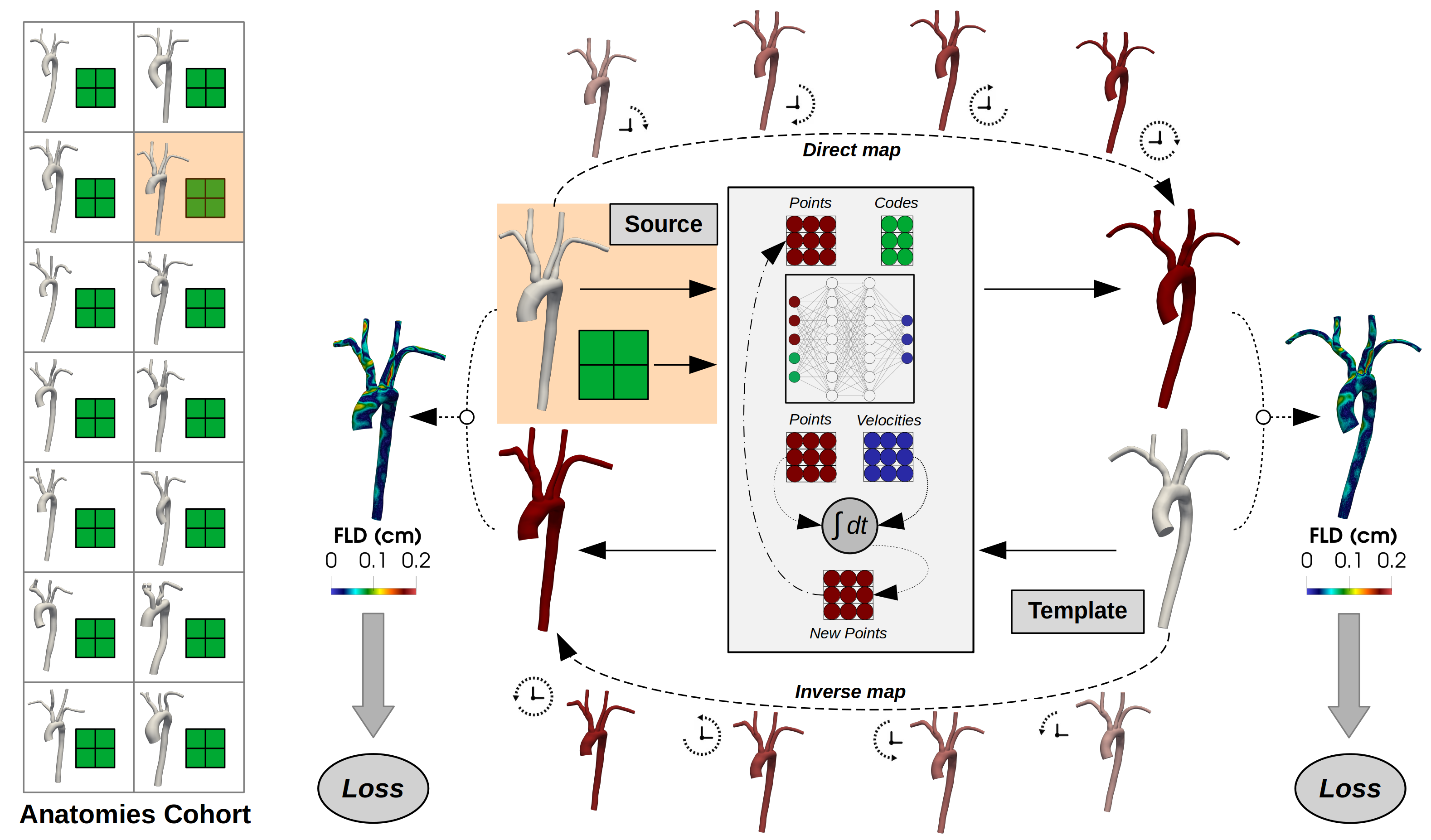}
\caption{General structure of the AD--SVFD model. The proposed approach leverages deep learning techniques to perform the diffeomorphic registration of vascular anatomies to a reference. Invertible ambient space deformations are modeled as solutions at unit time of ODEs, whose right--hand sides are parametrized by neural networks. The source and template geometries, represented as point clouds, are provided as input to AD--SVFD. The direct (top part of the image) and inverse (bottom part of the image) transforms are obtained by integrating the flow equations forward and backward in time, respectively. Geodesic paths can be visualized by morphing the input shapes at intermediate stages during the ODE integration. Generalization capabilities are enabled by associating each source shape with a trainable latent code (in green). The baseline model is optimized by minimizing the Chamfer distance (CD) between the mapped and the target geometries. Pointwise errors are quantified through the forward local distance (FLD), expressed in \si{\centi\metre}, namely the distance of each point in the mapped geometry from the closest one in the target.}
\label{fig: general model}
\end{figure}

In this work, we present a DL--based model for the deformable registration and synthetic generation of vascular anatomies, named AD--SVFD (\emph{Auto--Decoder Stationary Vector Field Diffeomorphism}). The general structure of AD--SVFD, reported in Figure~\ref{fig: general model}, is inspired to the models introduced by \emph{Amor et al.} in~\cite{amor2022resnet} (\emph{ResNet--LDDMM}), by \emph{Kong et al.} in~\cite{kong2024sdf4chd} (\emph{SDF4CHD}), and by \emph{Croquet et al.} in~\cite{croquet2021unsupervised}. Analogously to~\cite{amor2022resnet, croquet2021unsupervised}, AD--SVFD treats geometries as three--dimensional point clouds and employs \emph{ad hoc} data attachment measures to compensate for the absence of ground--truth point--to--point correspondences. The vascular shapes registration is achieved by deforming the ambient space according to an optimizable diffeomorphic map. The latter is approximated as the solution at unit time of an ordinary differential equation (ODE), whose time--independent right--hand side, representing a velocity field, is expressed through a fully--connected artificial neural network (ANN) (\emph{Neural ODE} paradigm \cite{chen2018neural}).  
Another major feature of AD--SVFD is its auto--decoder (AD) structure, introduced in a similar context in \cite{park2019deepsdf} (\emph{DeepSDF}) and then further exploited e.g. in \cite{kong2024sdf4chd}. In fact, AD--SVFD enables the simultaneous registration of a cohort of source shapes to a pre--defined common reference by introducing low--dimensional learnable latent codes, that are provided as input to the model and that condition its weights. As such, AD--SVFD configures as a self--conditional neural field~\cite{peng2009conditional}, since the conditioning variable is part of the model trainables. Compared to the more widely employed auto--encoders (AEs) \cite{kingma2013auto, higgins2016beta, kingma2019introduction}, that obtain latent input representations through a trainable encoding network, ADs entail faster and lighter optimization processes. Indeed, they roughly halve model complexity, at the cost of a cheap latent code inference procedure to be performed at the testing stage.
Other than featuring improved generalization capabilities and favoring efficient weight sharing, implicit neural representations through latent codes also enable generative AI applications~\cite{bond2021deep}. Indeed, synthetic anatomies can be crafted by drawing samples from empirical distributions, defined over the latent space, and by applying the associated inverse transforms to the reference geometry. \\

Let $\mathcal{T}$ denote the template (or reference) geometry and let $\{\mathcal{S}_i\}_{i=1}^{N^s}$ denote the cohort of available patient--specific vascular anatomies; the latter will be referred to as the \emph{source cohort} in the following. In particular, $\mathcal{T}$ and $\mathcal{S}_i$ identify three--dimensional closed surfaces, that are represented as weighted point clouds of the form:
\begin{equation}
\label{eq: point clouds}
\mathcal{T} := \left\{\left(\bm{x}_j^t, w_j^t\right)\right\}_{j=1}^{M^t}; \qquad \qquad
\mathcal{S}_i := \left\{\left(\bm{x}_{i,j}^s, w_{i,j}^s\right)\right\}_{j=1}^{M^s_i} \quad \text{for } i = 1, \cdots, N^s~.
\end{equation}
Here, $\bm{x}_j^t, \bm{x}_{i,j}^s \in \R^3$ are, respectively, the template and source points, and $w_j^t, w_{i,j}^s \in \R^+$ are the associated weights, which add up to one. In general, the weights associated with isolated points in the cloud should be large, while those in regions of high local density should be lower. In this work, we construct the point clouds from available triangular surface meshes by selecting the cell centers as points and computing the weights as the corresponding (normalized) cell areas.
To facilitate training, we perform a preliminary rigid registration of the source shapes to the template, based on the Coherent Point Drift algorithm~\cite{myronenko2010point}, and we apply an anisotropic rescaling. In this way, we embed every geometry in the unit cube $\Omega := [0,1]^3$ and we can assume that $\bm{x}_j^t, \bm{x}_{i,j}^s \in \Omega$ without loss of generality.
It is worth remarking that a tailored template shape can be estimated from the set of available anatomies, as done e.g. in \cite{durrleman2014morphometry, gori2017bayesian, yang2022implicitatlas, kong2024sdf4chd}. However, for simplicity, in this work we simply select one patient--specific anatomy to serve as a reference. \\

In mathematical terms, our goal is to find a set of diffeomorphisms $\{\vec{\varphi}_i\}_{i=1}^{N^s}$ that solve the following minimization problem:
\begin{equation}
\label{eq: optimization problem 1}
\left(\vec{\varphi}_1^*, \cdots, \vec{\varphi}_{N^s}^*\right) =
\argmin_{\left(\vec{\varphi}_1, \cdots, \vec{\varphi}_{N^s}\right)} \frac{1}{N^s} \sum_{i=1}^{N^s} \left(\mathcal{D}\left(\vec{\varphi}_i(\mathcal{S}_i), \mathcal{T}\right) + \mathcal{D}\left(\mathcal{S}_i, (\vec{\varphi}_i)^{-1}(\mathcal{T})\right)\right)~,
\end{equation}
where $\mathcal{D}: \R^{M_1} \times \R^{M_2} \to \R^+$ is some discrepancy measure between two point clouds, of cardinalities $M_1, M_2 \in \N$. Hence, we want to learn a family of invertible ambient space deformations, whose elements allow to optimally $(i)$ map the source shapes to the template via the direct transforms $\{\vec{\varphi}_i^*\}_{i=1}^{N^s}$ and $(ii)$ map the template shape to the sources via the inverse transforms $\{(\vec{\varphi}_i^*)^{-1}\}_{i=1}^{N^s}$. As discussed before, the AD--SVFD model features an auto--decoder structure, through the use of low--dimensional latent codes $\{\bm{z}_i\}_{i=1}^{N^s}, \ \bm{z}_i \in \R^{N_z}$, associated to the source shapes. In this way, the ambient space deformation map associated to $\mathcal{S}_i$ can be expressed as $\vec{\varphi}_i(\bm{x}) = \vec{\varphi}(\bm{x}; \bm{\Theta}, \bm{z}_i)$, so to entirely encapsulate the input dependency into the shape code. Hence, the optimization problem in Eq.\eqref{eq: optimization problem 1} can be conveniently rewritten as follows: find $\bm{\Theta}^* \in \R^{N_\Theta}, \ \bm{z}_i^* \in \R^{N_z} \ \text{for} \ i = 1, \cdots, N_s$, such that
\begin{equation*}
\label{eq: optimization problem 2}
\bm{\Theta}^*, \ \left(\bm{z}_1^*, \cdots, \bm{z}_{N^s}^*\right) =  \argmin_{\bm{\Theta}, \ \left(\bm{z}_1, \cdots, \bm{z}_{N^s}\right)} \frac{1}{N^s} \sum_{i=1}^{N^s} \mathcal{E}\Bigl(\mathcal{S}_i, \mathcal{T}, \vec{\varphi}(\ \cdot \ ; \bm{\Theta}, \bm{z}_i)\Bigr)~,
\end{equation*}
where $\mathcal{E}(\mathcal{S}, \mathcal{T}, \vec{\phi}) := \mathcal{D}(\vec{\varphi}(\mathcal{S}), \mathcal{T}) + \mathcal{D}(\vec{\varphi}^{-1}(\mathcal{T}), \mathcal{S})$ denotes the bidirectional mapping error between two point clouds $\mathcal{S}$ and $\mathcal{T}$, through the diffeomorphism $\vec{\varphi}$.\\

Adopting the stationary vector field (SVF) parametrization of diffeomorphisms \cite{ashburner2007fast, hernandez2009registration} (as done in \cite{croquet2021unsupervised, kong2024sdf4chd}), we exploit the Neural ODE paradigm~\cite{chen2018neural} to express the map $\varphi_i$ as the solution at unit time to the following learnable ODE:
\begin{equation}
\label{eq: SVF ODE}
\frac{\partial \vec{\varphi}_i(\bm{x}; t)}{\partial t} = \vec{v}\left(\vec{\varphi}_i(\bm{x}; t), \bm{\Theta}, \bm{z}_i\right) \quad \text{such that} \quad \vec{\varphi}_i(\bm{x}; 0) = \bm{x}~,
\end{equation}
where the vector $\bm{\Theta} \in \R^{N_{\Theta}}$ collects the trainable parameters of an ANN. As demonstrated in \cite{ma2022cortexode}, if the ANN that expresses the velocity field $\vec{v}$ is fully--connected and features \emph{ReLU} or \emph{Leaky--ReLU} activation functions, then $\vec{v}$ is Lipschitz continuous and Eq.\eqref{eq: SVF ODE} admits a unique solution. Consequently, the inverse transform $(\vec{\varphi}_i)^{-1}$, that deforms the ambient space so as to overlap the template point cloud $\mathcal{T}$ to the source one $\mathcal{S}_i$, can be found by integrating Eq.\eqref{eq: SVF ODE} backward in time. In this work, we employed the first--order forward Euler and modified Euler schemes to numerically integrate the diffeomorphic flow equations forward and backward in time, respectively, considering $K=10$ discrete time steps, as in~\cite{amor2022resnet}. \\

Our approach is developed under the assumption that all shapes share the same topology. Conversely, it is not possible to guarantee the existence (and uniqueness) of a diffeomorphic flow field that exactly deforms one into the other. In fact, non--rigid registration under topological variability remains an open challenge~\cite{sahilliouglu2020recent}. \\ 

To train the AD--SVFD model, we employ the following loss function:
\begin{equation}
\label{eq: loss function}
\mathcal{L} (\bm{\Theta}, \bm{Z}) := \frac{1}{N^s} \sum_{i=1}^{N^s} \Bigl( \mathcal{E}\bigl(\mathcal{S}_i, \mathcal{T}; \vec\varphi(\cdot;\bm{\Theta}, \bm{z}_i)\bigr)\Bigr) + w_z \Norm{\bm{Z}}_2^2 + w_\Theta \Norm{\bm{\Theta}}_2^2 + w_v \ \mathcal{L}_{\operatorname{reg}}(\bm{\Theta})~, 
\end{equation}
where $w_z, w_\Theta, w_v \in \R^+$ are scalar weight factors, $\bm{Z} \in \R^{N_z \times N_s}$ is a matrix collecting the shape codes associated to the $N_s$ training shapes, and $\mathcal{L}_{\operatorname{reg}}$ is a regularization term that constrains the velocity field learned by the ANN. In the numerical experiments, we explore multiple alternatives for the data attachment measure $\mathcal{D}$ that appears in the definition of the bidirectional mapping error $\mathcal{E}$. Specifically, we consider the Chamfer Distance (CD)~\cite{feydy2017optimal}, the point--to--plane Chamfer Distance (PCD)~\cite{tian2017geometric}, the Chamfer Distance endowed with a penalization on the normals' orientation scaled by the factor $w_n \in \R^+$ (denoted as NCD), and the debiased Sinkhorn divergence (SD)~\cite{cuturi2013sinkhorn}. Furthermore, we exploit the availability of weights (see Eq.\eqref{eq: point clouds}) to derive data adherence measures, that should be better able to deal with unevenly distributed point clouds. 
The training procedure is carried out with the \emph{Adam} optimizer~\cite{kingma2014adam}, considering $E=500$ epochs, a batch size $B=8$, and setting the same learning rate $\lambda \in \R^+$ to update the ANN parameters and the shape codes. At each epoch, sub--clouds made of $M=2,000$ points are adaptively sampled to limit computational efforts and memory requirements. More details on both the data attachment measures and the training pipeline are provided in \emph{Methods}.
During testing, we can combine \emph{Adam} with higher--order memory--intensive methods, such as L--BFGS~\cite{bottou2018optimization}, since only the latent code entries have to be optimized. Specifically, we first run $100$ epochs using \emph{Adam} and then we fine--tune the predictions using L--BFGS for $10$ epochs. Additionally, preliminary numerical results suggested using a learning rate $50$ times larger than the one employed for training with \emph{Adam}, as this facilitates and speeds--up convergence.


\section*{Results}  \label{sec: results}
\begin{figure}[t!]
\centering
\includegraphics[width=0.99\textwidth]{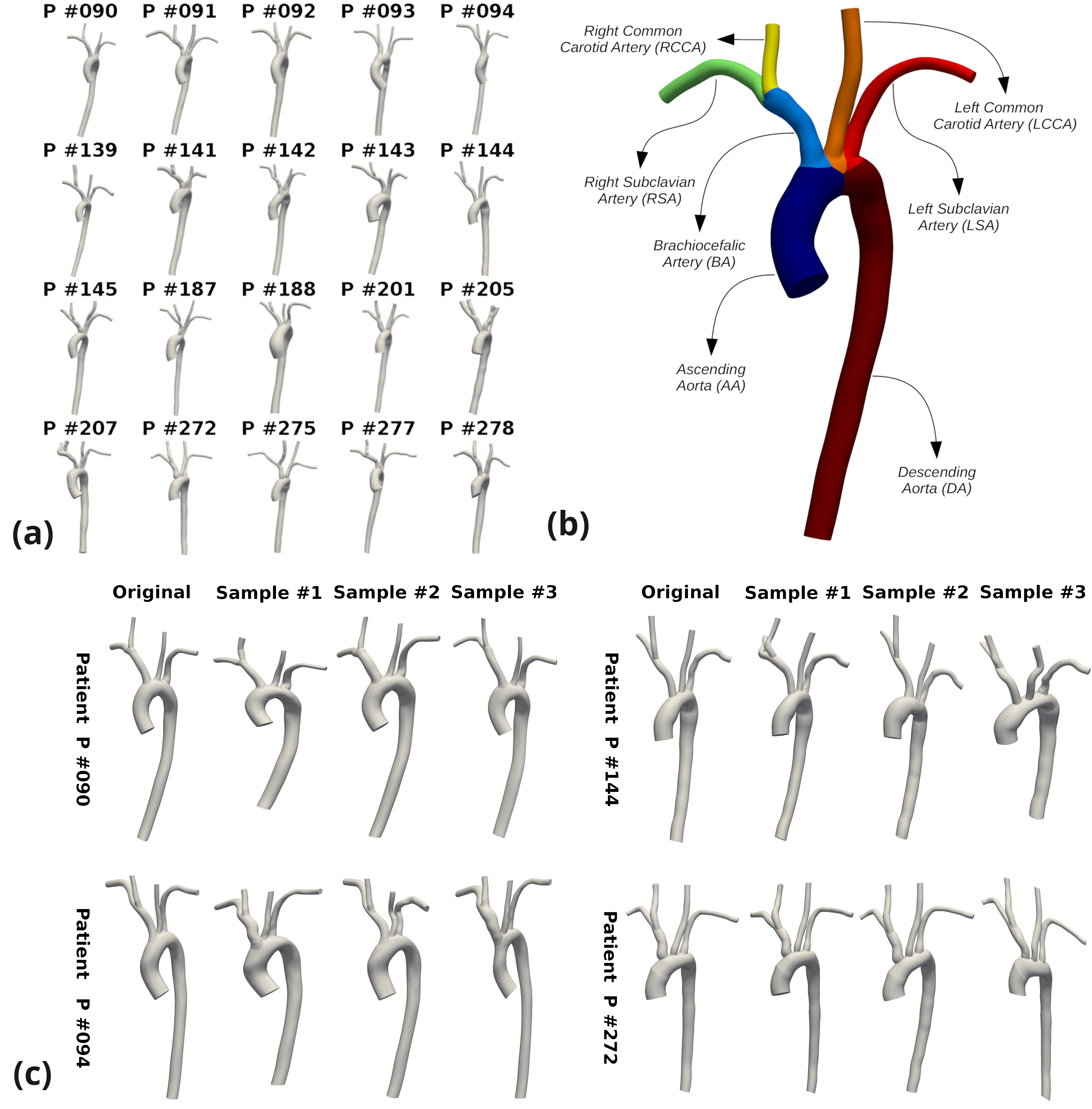}
\caption{Healthy aortic shapes dataset overview. In particular: (a) original dataset of patient--specific anatomies; (b) topology of the considered geometries, with nomenclature of the different branches; (c) original shape and three synthetic samples, generated by deforming four anatomies with the implemented data augmentation pipeline.}
\label{fig: dataset}
\end{figure}

We present the numerical experiments conducted on the AD--SVFD model and briefly discuss the obtained results. All tests have been performed starting from a dataset containing $20$ healthy aortic anatomies, that have been segmented from medical images (CT--scans and MRIs) using \emph{SimVascular}~\cite{updegrove2017simvascular} (see Figure~\ref{fig: dataset}~(a)) and are publicly available in the \emph{Vascular Model Repository}~\cite{wilson2013vascular}. As depicted in Figure~\ref{fig: dataset}~(b), we underline that all the geometries share the same topology, that comprises the aortic vessel (ascending chunk (AA) and descending chunk (DA)), the brachiocefalic artery (BA), the left and right subclavian arteries (LSA, RSA), and the left and right common carotid arteries (LCCA, RCCA). To generate weighted point cloud representations of the shapes, we created volumetric tetrahedral computational meshes and extracted triangulations of the external surfaces. This allowed us to choose the surface cell centers as the cloud points, and the surface cell areas as the associated weights (see Eq.\eqref{eq: point clouds}). \\ 

The number of available anatomies is evidently too low to train a DL--based model, whose performances drastically depend on the amount of data at disposal. Therefore, we implemented an \emph{ad hoc} data augmentation pipeline, based on Coherent Point Drift (CPD) rigid registration~\cite{myronenko2010point} and thin--plate spline (TPS) interpolation~\cite{duchon1977splines}. We refer to \emph{Appendix}~\ref{app: data augmentation} for a detailed description. This allowed us to assemble a dataset made of $902$ anatomies, out of which $882$ have been artificially generated. A few synthetic anatomies are reported in Figure~\ref{fig: dataset}~(c). We perform a train--test splitting, reserving $38$ shapes solely for testing. In particular, $2$ of the test geometries belong to the original dataset, and their corresponding augmented versions are not taken into account for training; the remaining $36$ test geometries are instead augmented versions of the $18$ original anatomies included in the training dataset ($2$ augmented shapes per patient). Except for the cross--validation procedure, all the numerical tests are carried out considering the same training and testing datasets. They have been obtained by reserving patients \emph{P\#093} and \emph{P\#278} for testing, which results in employing $780$ geometries for training. We remark that patient \emph{P\#091} serves as a reference in all test cases. \\

Most hyperparameters of the ANN model have been calibrated in a simplified single shape--to--shape registration scenario, using the Tree--structured Parzen Estimator (TPE) Bayesian algorithm~\cite{bergstra2011algorithms, bergstra2013making}. We refer to \emph{Appendix}~\ref{app: TPE tuning} for a complete list of the hyperparameters and for a detailed description of the tuning procedure. Besides dictating the specifics of the ANN model architecture, the calibration results suggested to set the learning rate $\lambda_\Theta = \lambda_z = \lambda = 10^{-3}$, and the loss weights $w_v = 10^{-4}$ and $w_z = 10^{-3}$ (see Eq.\eqref{eq: loss function}). Unless differently specified, the loss is computed considering the standard (i.e. not weighted) CD as a data attachment measure. The model accuracy is quantified through the forward and backward local distances (FLD and BLD), expressed in \si{\centi\metre}. The former identifies the distance of each point in the mapped geometry from the closest one in the target, while the latter is the distance of each point in the target from the closest one in the mapped geometry. \\

All computations were performed on the \href{https://www.sherlock.stanford.edu/}{\emph{Sherlock}} cluster at \emph{Stanford University}, employing an AMD 7502P processor (32 cores), 256 GB RAM, HDR InfiniBand interconnect, and a single NVIDIA GeForce RTX 2080 Ti GPU. We note that the exact reproducibility of the results cannot be guaranteed, owing to the use of non--deterministic algorithms provided by the \emph{PyTorch} library to enhance efficiency.

\subsection*{Test 1: Latent shape codes}  

\begin{figure}[!t]
\centering
\includegraphics[width=.99\textwidth]{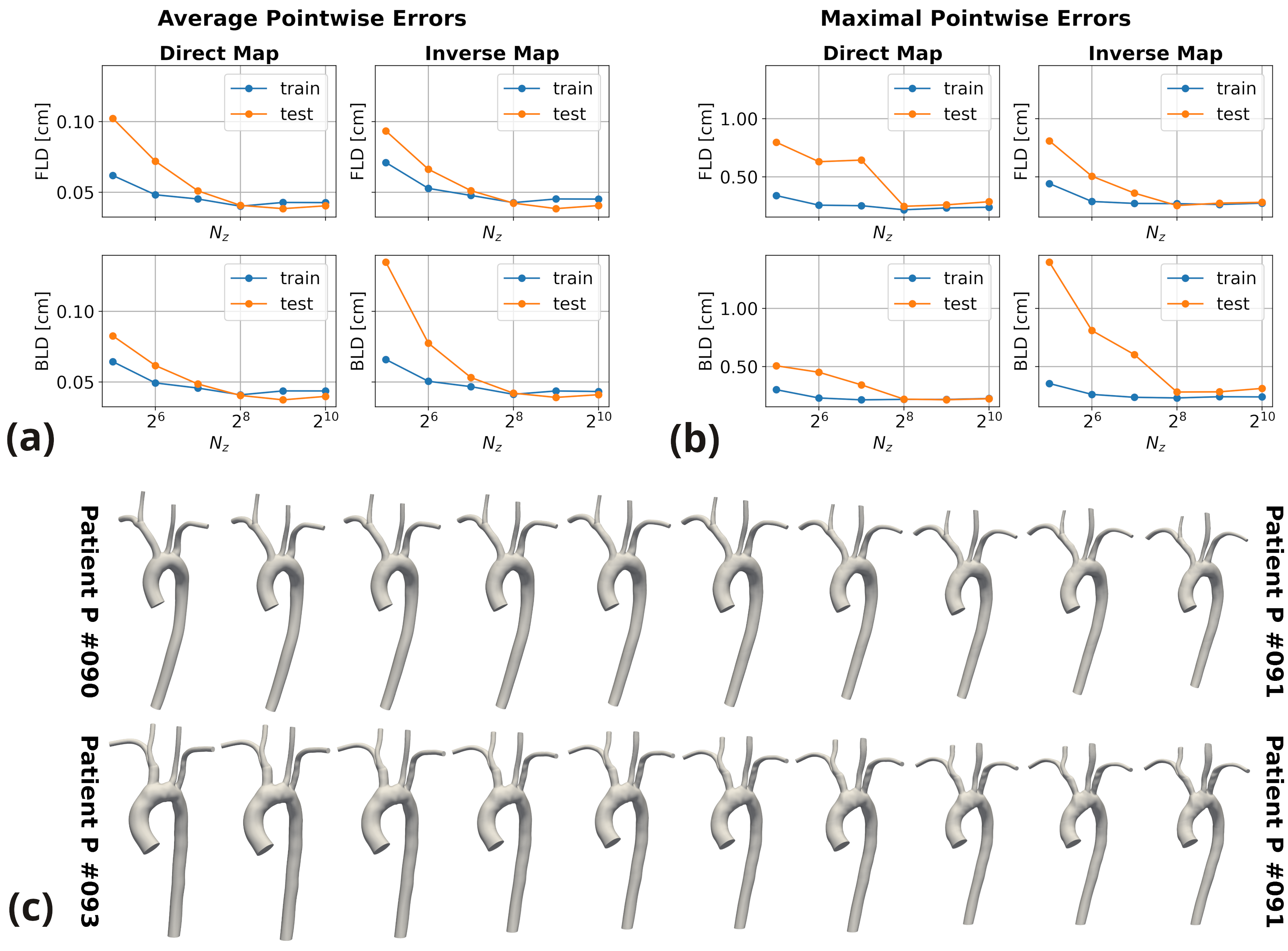}
\caption{Deformable mapping results of the baseline AD--SVFD model. In particular, we report the average (a) and maximal (b) pointwise errors --- quantified through the forward and backward local distances FLD and BLD, in $\si{\centi\metre}$ --- on training and testing datapoints, obtained for different shape code dimensions $N_z$; in (c), we show the geodesic paths between two source shapes (P\#090 for training, P\#093 for testing) and the reference shape (P\#091), generated by numerical integration of the diffeomorphic flow equations by the forward Euler method at $K=10$ intermediate steps.}
\label{fig: zs error trends}
\end{figure}

We investigate the effect of shape codes on the AD--SVFD model results, focusing in particular on the latent space dimension $N_z$. We point out that the training errors are computed only on the $18$ original shapes. \\

Figure~\ref{fig: zs error trends} reports the average (a) and maximal (b) FLD and BLD for both the direct and the inverse deformation, considering different values of $N_z$. On the one hand, the results demonstrate that the latent space dimension should be taken sufficiently large, in order to effectively condition the model weights towards accurate approximations of the diffeomorphic maps. On the other hand, we notice that model accuracy stalls for large values of $N_z$, suggesting redundant information in the shape codes. Ultimately, we select $N_z = 256$ as the latent dimension, since it appears to optimally balance accuracy and efficiency. In terms of generalization power, we note that training and testing errors are comparable for $N_z \geq 256$, thus indicating that no overfitting phenomenon occurs. Incidentally, we remark that no major discrepancy between the registration errors on original and augmented testing geometries can be observed. For instance, only marginally lower maximal FLD are obtained on the augmented geometries, both considering the direct and the inverse map (direct map errors: $0.2774 \ \si{\centi\metre}$ vs. $0.2822 \ \si{\centi\metre}$; inverse map errors: $0.2562 \ \si{\centi\metre}$ vs. $0.2613 \ \si{\centi\metre}$).
In Figure~\ref{fig: zs error trends}~(c), we appreciate how AD--SVFD smoothly and gradually warps the source shapes into the reference one, through the forward--in--time numerical integration of the learnable diffeomorphic flow equations (see Eq.\eqref{eq: SVF ODE}) by the explicit Euler method.

\subsection*{Test 2: Data attachment measures}

\begin{table}[!t]
	\caption{Registration results of AD--SVFD considering different data attachment measures. In particular, we report the maximal pointwise errors on training and testing datapoints, obtained for six different data adherence metrics. The errors are quantified through the forward and backward local distances (FLD and BLD), expressed in \si{\centi\metre}. The best value for each performance metric is marked in green. 
    For reference, the template shape inlet diameter is $1.31$ \si{\centi\metre}, while the average inlet diameter in the dataset is $1.45$ \si{\centi\metre}. 
    Acronyms. CD: Chamfer distance; PCD: point--to--surface Chamfer distance~\cite{tian2017geometric}; NCD: Chamfer distance with normals penalization; SD: debiased Sinkhorn divergence~\cite{cuturi2013sinkhorn}. Notation: the $W$ superscript denotes the use of a weighted measure.}

	\label{tab: loss function results}
	\def\arraystretch{1.25}
	\resizebox{\textwidth}{!}{
    \begin{tabular}{C{1.5cm} C{1.35cm} C{1.35cm} C{1.35cm} C{1.35cm} C{0.01cm} C{1.35cm} C{1.35cm} C{1.35cm} C{1.35cm}}
    \toprule
    &
    \multicolumn{4}{c}{\large\textbf{Train Errors (in cm)}}&&
    \multicolumn{4}{c}{\large\textbf{Test Errors (in cm)}}\\
    
    \cmidrule(lr){2-5} \cmidrule(lr){7-10}
    
    &
    \multicolumn{2}{c}{\large\textbf{Direct}}&
    \multicolumn{2}{c}{\large\textbf{Inverse}}&&
    \multicolumn{2}{c}{\large\textbf{Direct}}&
    \multicolumn{2}{c}{\large\textbf{Inverse}} \\ 
    
    \cmidrule(lr){2-3} \cmidrule(lr){4-5} 
    \cmidrule(lr){7-8} \cmidrule(lr){9-10} 
    
    \large\textbf{Loss} 
    & \textbf{FLD} & \textbf{BLD} & \textbf{FLD} & \textbf{BLD} &
    & \textbf{FLD} & \textbf{BLD} & \textbf{FLD} & \textbf{BLD} \\
    
    \midrule
    
    $\bm{\mathcal{D}}_{CD}$ & 0.2162 & 0.2175 & 0.2686 & \cellcolor{green!25}0.2297 && \cellcolor{green!25}0.2777 & \cellcolor{green!25}0.2253 & 0.2562 & \cellcolor{green!25}0.2642 \\
    $\bm{\mathcal{D}}_{CD}^W$ & 0.2412 & 0.2497 & 0.2869 & 0.2564 &&  0.4088 & 0.2479 & 0.2952 & 0.4283 \\
    $\bm{\mathcal{D}}_{PCD}$ & 0.3195 & \cellcolor{green!25}0.2165 & 0.3225 & 0.2611 &&  0.3138 & 0.2516 & 0.2749 & 0.3540 \\
    $\bm{\mathcal{D}}_{PCD}^W$ & 0.2515 & 0.2510 & 0.2958 & 0.2579 && 0.3489 & 0.2439 & 0.3043 & 0.3686  \\

    $\bm{\mathcal{D}}_{NCD}$ & \cellcolor{green!25}0.2033 & 0.2166 & \cellcolor{green!25}0.2628 & 0.2396 && 0.2965 & 0.2260 & \cellcolor{green!25}0.2497 & 0.3090 \\
    
    \midrule
    
    $\bm{\mathcal{D}}_{SD}$ & 0.4887 & 0.3795 & 0.5355 & 0.3923 && 0.4519 & 0.4695 & 1.1094 & 0.4384 \\
    $\bm{\mathcal{D}}_{SD}^W$ & 0.5866 & 0.3861 & 0.5155 & 0.3917 && 0.4156 & 0.4155 & 0.8275 & 0.4420 \\
    
    \bottomrule
    
    \end{tabular}
}
\end{table}

\begin{figure}[t!]
	\centering
	\includegraphics[width=0.95\textwidth]{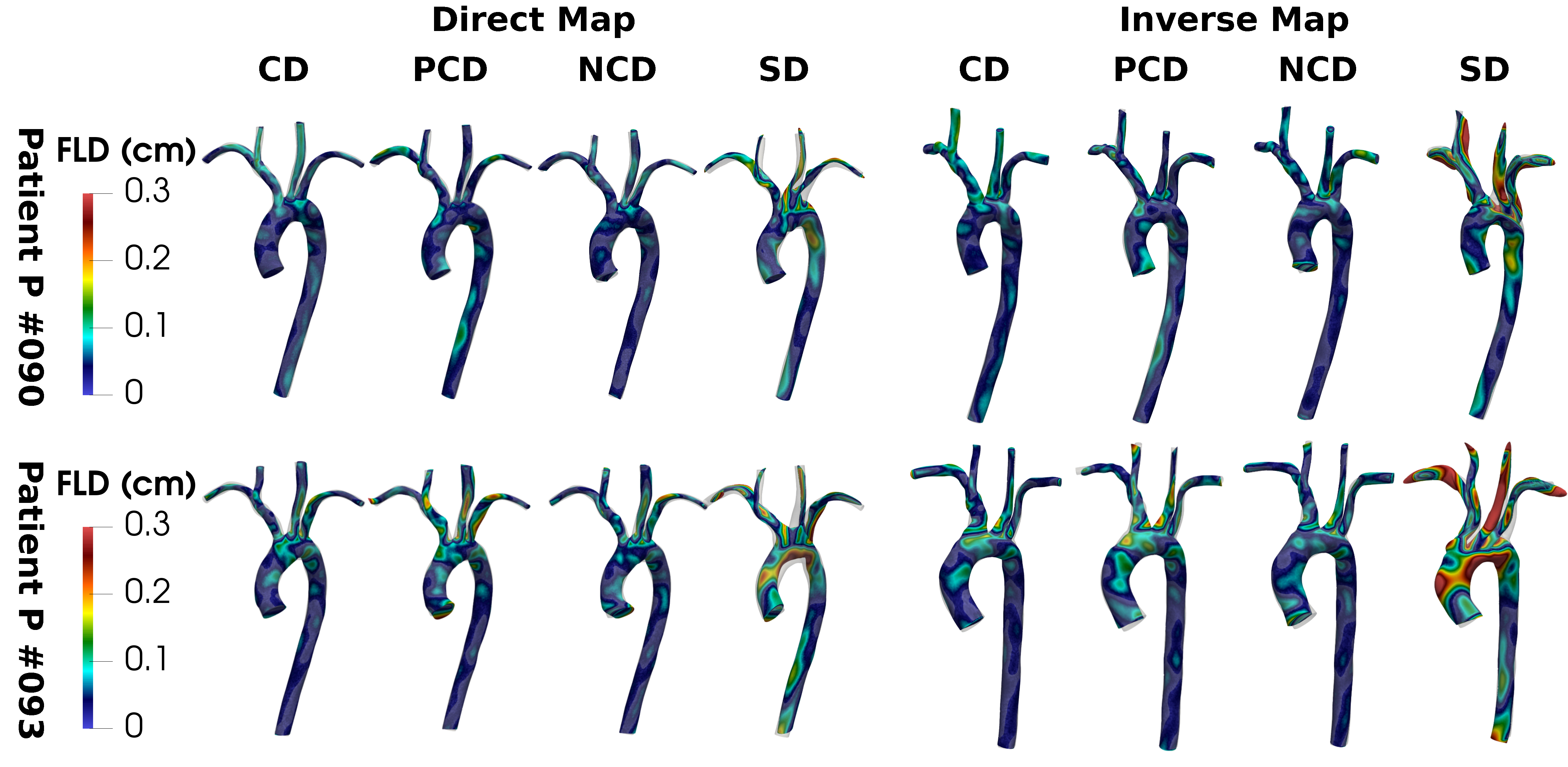}
	\caption{Registration results of AD--SVFD considering different data attachment measures. In particular, we show the direct and inverse mapping pointwise errors, obtained on a training (\emph{P\#090}) and a testing (\emph{P\#093}) datapoint, for four different data adherence metrics. The errors are quantified through the forward local distance (FLD), expressed in \si{\centi\metre}, namely the distance of each point in the mapped geometry from the closest one in the target. 
    For reference, the inlet diameters are $1.31$ \si{\centi\metre} for the template shape, $1.21$ \si{\centi\metre} for \emph{P\#090}, and $1.32$ \si{\centi\metre} for \emph{P\#093}. 
    Acronyms. CD: Chamfer distance; PCD: point--to--surface Chamfer distance~\cite{tian2017geometric}; NCD: Chamfer distance with normals penalization; SD: debiased Sinkhorn divergence~\cite{cuturi2013sinkhorn}.}
	\label{fig: data attachment results}
\end{figure}

We analyse the AD--SVFD model performances considering the different data attachment measures mentioned in the \emph{Introduction}. We refer to \emph{Methods} for a detailed description of the different options and of their specifics. Table~\ref{tab: loss function results} reports the maximal pointwise errors on both training and testing datapoints. To quantitatively compare the results, we evaluate the maximal pointwise FLD and BLD, even when metrics different from the (unweighted) CD are used in the loss. Since this approach may introduce a bias in the analysis, we also provide a qualitative accuracy assessment through Figure~\ref{fig: data attachment results}. \\

From both a quantitative and a qualitative standpoint, the best results are obtained considering the baseline model, which employs unweighted CD as a data attachment measure. Indeed, this model yields precise geometry reconstructions on both training and testing shapes, and it is associated with the lowest training time (equal to $7h40m$) and with an average testing time of just $1m28s$ per shape\footnote{Average testing times have been computed on the \href{https://www.epfl.ch/research/facilities/scitas/hardware/kuma/}{\emph{Kuma}} cluster at \emph{EPFL}, considering a single NVIDIA H100 SXM5 GPUs, 94 GB RAM (HBM2e), memory bandwidth of 2.4 TB/s, Interconnected with NVLink, 900 GB/s bandwidth.}. Incorporating a penalization of the normals' orientation in the loss (with $w_n = 10^{-2}$) allows for marginal accuracy improvements, but at the cost of a much larger training time (equal to $16h07m$), due to the increased number of model evaluations. Incidentally, the larger memory requirements induced by the normals' calculation prevent the use of L--BFGS at inference. To mitigate this issue, we replace NCD with unweighted CD at testing; this allows to retain acceptable accuracy levels, even though worse than the training ones, at equivalent inference times. Neither leveraging the weights associated with the point clouds nor adopting PCD improves the mapping quality; in fact, both approaches are substantially outperformed by the baseline model. With a specific focus on PCD, from Figure~\ref{fig: data attachment results} we can observe marked discrepancies at the upper branches, which in the case of patient \emph{P\#093} tend to squeeze into unrealistic flat morphologies. 
Lastly, we remark that the registration quality gets considerably worse when using debiased SD. Indeed, the deformed geometries take unlikely convoluted shapes, which become twisted and almost flat in the upper branches region. From a quantitative point of view, this translates into errors that roughly double the ones obtained with CD. Furthermore, compared to the baseline model, the heavier costs associated with the calculation of SD entail drastic increases in the durations of both training (from $7h40m$ to $27h15m$) and testing (from $1m28s$ to $3m08s$ per shape on average).

\subsection*{Test 3: Comparison with state--of--the--art methods}

\begin{table}[!t]
\caption{Comparison test of AD--SVFD with six alternative registration methods. In particular, we report the average and maximal pointwise errors on patients \emph{P\#090} and \emph{P\#278}, obtained considering CPD~\cite{myronenko2010point}, TPS~\cite{duchon1977splines}, LDDMM~\cite{bone2018deformetrica}, ResNet--LDDMM~\cite{amor2022resnet} (optionally endowed with a penalty of the inverse deformation, I--ResNet--LDDMM), SDF4CHD~\cite{kong2024sdf4chd} and AD--SVFD. The errors are quantified through the forward and backward local distances (FLD and BLD), expressed in \si{\centi\metre}. For reference, the inlet diameters are: $1.31$ \si{\centi\metre} for \emph{P\#091} (template); $1.22$ \si{\centi\metre} for \emph{P\#090}; $1.52$ \si{\centi\metre} for \emph{P\#278}.}

\label{tab: results comparison}
\def\arraystretch{1.25}
\resizebox{\textwidth}{!}{
\begin{tabular}{C{0.4cm} L{1.7cm} C{1.35cm} C{1.35cm} C{1.35cm} C{1.35cm} C{1.35cm} C{1.35cm} C{1.35cm} C{1.35cm}}
	\toprule
	& &
	\multicolumn{4}{c}{\large\textbf{Max Errors (in cm)}}&
	\multicolumn{4}{c}{\large\textbf{Avg Errors (in cm)}}\\
	
	\cmidrule(lr){3-6} \cmidrule(lr){7-10}
	
	& &
	\multicolumn{2}{c}{\large\textbf{Direct}}&
	\multicolumn{2}{c}{\large\textbf{Inverse}}&
	\multicolumn{2}{c}{\large\textbf{Direct}}&
	\multicolumn{2}{c}{\large\textbf{Inverse}}\\ 
	
	\cmidrule(lr){3-4} \cmidrule(lr){5-6} \cmidrule(lr){7-8} \cmidrule(lr){9-10}
	
	& \textbf{Method}
	& \textbf{FLD} & \textbf{BLD} & \textbf{FLD} & \textbf{BLD}
	& \textbf{FLD} & \textbf{BLD} & \textbf{FLD} & \textbf{BLD} \\
	
	\midrule
	
	\multirow{8}{*}{\rotatebox[origin=c]{90}{\centering \textbf{\large P\#090}}} &
	\textbf{CPD}& 1.4321 & 2.0983 & 1.3985 & 2.9714 & 0.2709 & 0.3351 & 0.3768 & 0.4669 \\
	& \textbf{TPS}        & 0.2918 & 0.2615 & 0.4305 & 0.3674 & 0.0777 & 0.0770 & 0.1050 & 0.1029 \\
	& \textbf{LDDMM}      & 0.1813 & \cellcolor{green!25}0.1227 & 0.2625 & 0.3332 & \cellcolor{green!25}0.0315 & \cellcolor{green!25}0.0290 & 0.0438 & 0.0412 \\
	\cmidrule(lr){2-10}
	& \textbf{ResNet}     & 0.2470 & 0.2806 & 0.4393 & 0.6520 & 0.0530 & 0.0554 & 0.0948 & 0.0983 \\
	& \textbf{I--ResNet}  & 0.1948 & 0.2269 & 0.2139 & 0.2466 & 0.0454 & 0.0468 & 0.0484 & 0.0469 \\
	\cmidrule(lr){2-10}
	& \textbf{SDF4CHD}    & 0.3861 & 1.7831 & 0.6250 & 0.4207 & 0.0399 & 0.0684 & 0.0502 & \cellcolor{green!25}0.0405 \\
	& \textbf{AD--SVFD}   & \cellcolor{green!25}0.1693 & 0.1923 & \cellcolor{green!25}0.2071 & \cellcolor{green!25}0.1719 & 0.0387 & 0.0399 & \cellcolor{green!25}0.0430 & 0.0411 \\
	
	\midrule
	
	\multirow{8}{*}{\rotatebox[origin=c]{90}{\centering \textbf{\large P\#278}}} &
	\textbf{CPD} & 1.5265 & 1.1772 & 1.3802 & 2.0471 & 0.2997 & 0.2962 & 0.3481 & 0.3648 \\
	& \textbf{TPS}       & 0.5092 & 0.3521 & 0.7974 & 0.7104 & 0.0649 & 0.0644 & 0.0987 & 0.0931 \\
	& \textbf{LDDMM}     & \cellcolor{green!25}0.1281 & \cellcolor{green!25}0.1681 & 0.5160 & 0.5441 & \cellcolor{green!25}0.0294 & \cellcolor{green!25}0.0285 & 0.0552 & 0.0692 \\
	\cmidrule(lr){2-10}
	& \textbf{ResNet}    & 0.3085 & 0.2720 & 0.3546 & 0.3594 & 0.0551 & 0.0555 & 0.0651 & 0.0615 \\
	& \textbf{I--ResNet} & 0.2805 & 0.2498 & 0.2986 & 0.3367 & 0.0486 & 0.0485 & 0.0547 & 0.0525 \\
	\cmidrule(lr){2-10}
	& \textbf{SDF4CHD}   & 0.2598 & 1.2918 & 1.0277 & \cellcolor{green!25}0.2754 & 0.0421 & 0.0604 & 0.0592 & 0.0464 \\
	& \textbf{AD--SVFD}  & 0.2166 & 0.1807 & \cellcolor{green!25}0.2817 & 0.2933 & 0.0379 & 0.0370 & \cellcolor{green!25}0.0420 & \cellcolor{green!25}0.0419  \\
	
	\bottomrule
	
\end{tabular} 
}	
\end{table}

\begin{figure}[t!]
\centering
\includegraphics[width=0.975\textwidth]{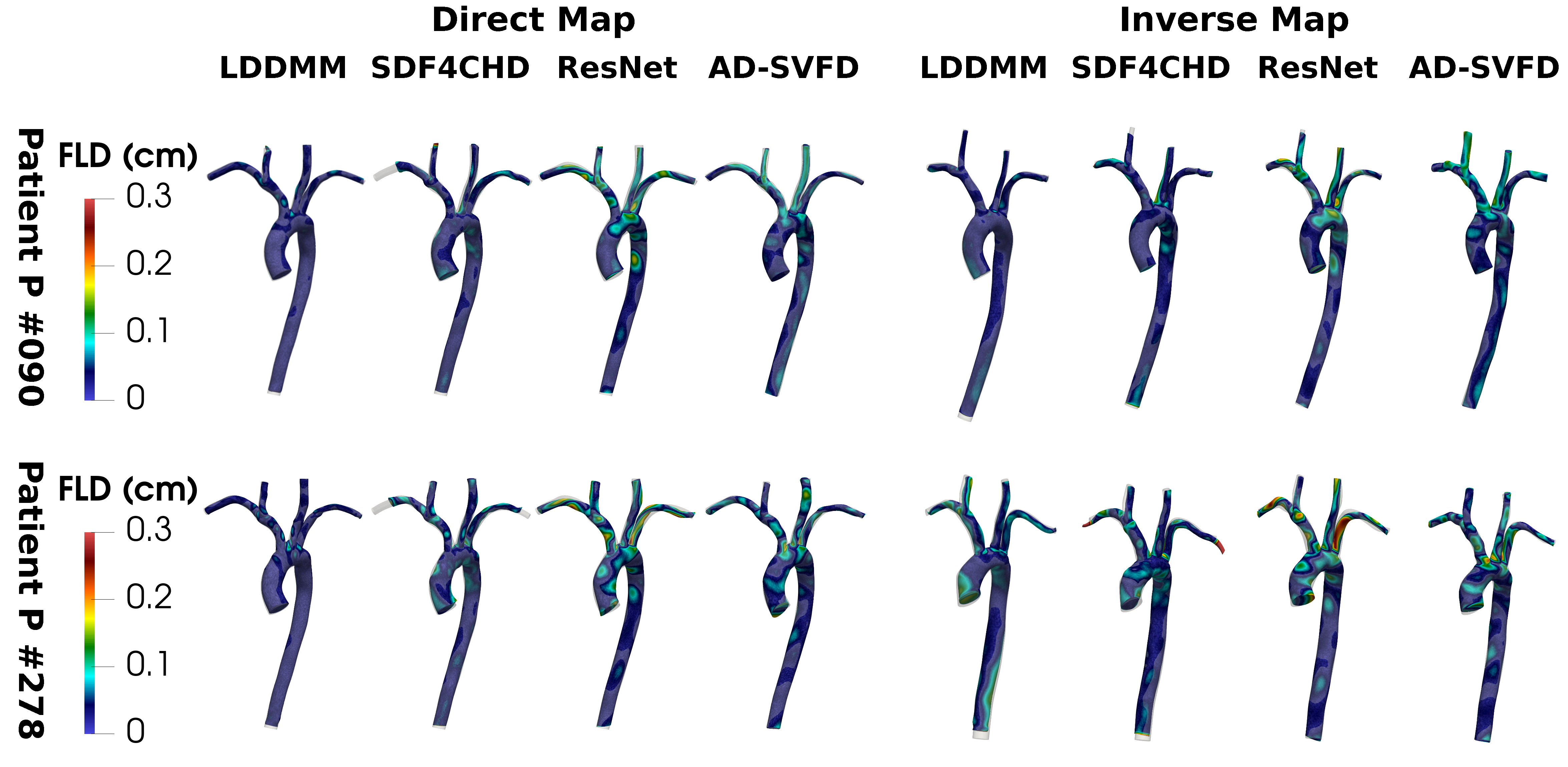}
\captionof{figure}{Registration results obtained with AD--SVFD and three alternative approaches. In particular, we show the direct and inverse mapping pointwise errors, obtained with LDDMM~\cite{bone2018deformetrica}, SDF4CHD~\cite{kong2024sdf4chd}, ResNet--LDDMM~\cite{amor2022resnet} and AD--SVFD on a training (\emph{P\#090}) and a testing (\emph{P\#278}) datapoint. The errors are quantified through the forward local distance (FLD), expressed in \si{\centi\metre}, namely the distance of each point in the mapped geometry from the closest one in the target. For reference, the inlet diameters are $1.31$ \si{\centi\metre} for the template shape, $1.21$ \si{\centi\metre} for \emph{P\#090}, and $1.52$ \si{\centi\metre} for \emph{P\#278}.}
\label{fig: comparison results}
\end{figure}
  
To fairly assess the capabilities of AD--SVFD, we run a comparison test with five alternative shape registration models. Specifically, we evaluate the mapping quality considering two different source shapes: \emph{P\#090} (training) and \emph{P\#278} (testing). We investigate the following models: Coherent Point Drift (CPD)~\cite{myronenko2010point} (rigid registration model, serving as a baseline), thin--plate spline (TPS) interpolation~\cite{duchon1977splines} (see \emph{Appendix}~\ref{app: data augmentation} for details), LDDMM~\cite{bone2018deformetrica}, SDF4CHD~\cite{kong2024sdf4chd} and ResNet--LDDMM~\cite{amor2022resnet}. A few aspects deserve attention.
\begin{itemize}
\item Except from SDF4CHD, all the other approaches perform single shape--to--shape registrations, without leveraging any form of implicit geometry representation. Hence, for these models there is no distinction between training and testing shapes. 
\item Using CPD, TPS and LDDMM, we can only estimate a one--directional map, warping the source shape into the reference one or viceversa. Therefore, the direct and inverse maps are retrieved by running two independent optimization processes. While this approach may improve registration accuracy, it comes at the cost of increased computational efforts and does not guarantee that the two maps compose to the identity.
\item As reported in \cite{amor2022resnet}, despite learning a diffeomorphism between two shapes, the ResNet--LDDMM model is solely optimized considering the source--to--template map result. To enhance inverse mapping quality, we introduce the \emph{I--ResNet--LDDMM} model. Compared to the baseline, the latter solves a multi--objective optimization problem, including both direct and inverse mapping results within the loss. To provide a fair comparison with AD--SVFD, we employ CD as a data attachment measure, rely on the modified Euler scheme to integrate the diffeomorphic flow ODE backward in time, and equally weigh direct and inverse errors.
\end{itemize}

Regarding the models' specifics, for ResNet--LDDMM and SDF4CHD we employ the ``optimal'' model structures and hyperparameter sets, as identified in \cite{amor2022resnet} and \cite{kong2024sdf4chd}, respectively. Furthermore, with SDF4CHD we do not exploit the DeepSDF model~\cite{park2019deepsdf} to learn the SDF representation of an \emph{Atlas} shape; instead, we use the pre--computed SDF of patient \emph{P\#091} to serve as reference. For LDDMM, we rely on the \emph{Deformetrica}~\cite{bone2018deformetrica} software, and perform single shape--to--shape registration employing the varifold distance, with a Gaussian kernel of width $0.8$ as data attachment measure. This last value, which leads to the optimization of roughly $1,000$ control points and momenta vectors, has been manually calibrated to balance efficiency and accuracy. \\

Table~\ref{tab: results comparison} reports the maximal and average pointwise errors of the direct and inverse mappings obtained on the two considered source shapes, for the different registration models. Figure~\ref{fig: comparison results} displays the results for four of the models. In summary, we can claim that AD--SVFD and LDDMM significantly outperform all the other approaches. In particular, LDDMM yields the most precise approximations of the direct map; however, its performances deteriorate and fall behind the ones of AD--SVFD on the inverse map, particularly because of discrepancies at the inlet/outlet faces. A similar consideration holds for the SDF4CHD model, that is capable of producing anatomies that closely match the target ones, but that often feature artifacts and/or completely miss the final portion of the smallest branches. In contrast with the results reported in \cite{amor2022resnet}, the residual neural network structure of ResNet--LDDMM does allow it to outperform the canonical LDDMM method. Nonetheless, we acknowledge that fine--tuning the main model hyperparameters to the present test case could sensibly improve the results. Additionally, we underline that the introduction of a penalty on the inverse mapping in ResNet--LDDMM determines minor but tangible improvements on all metrics.     

\subsection*{Test 4: Robustness assessment} 

\begin{table}[!t]
\caption{Cross--validation procedure results. In particular, we report the maximal pointwise errors on training and testing datapoints, solely considering original anatomies, obtained with the AD--SVFD model for the ten different folds during cross--validation. The errors are quantified through the forward and backward local distances (FLD and BLD), expressed in \si{\centi\metre}. The reported results are averages, that stem from three independent training procedures, conducted by setting different random seeds. For reference, the template shape inlet diameter is $1.31$ \si{\centi\metre}, while the average inlet diameter in the dataset is $1.45$ \si{\centi\metre}.}
\label{tab: results CV}
\def\arraystretch{1.25}
\resizebox{\textwidth}{!}{
\begin{tabular}{C{1.35cm} C{1.5cm} C{1.35cm} C{1.35cm} C{1.35cm} C{1.35cm} C{1.35cm} C{1.35cm} C{1.35cm} C{1.35cm} C{1.35cm}}
	\toprule
	& &
	\multicolumn{4}{c}{\large\textbf{Train Errors (in cm)}}&
	\multicolumn{4}{c}{\large\textbf{Test Errors (in cm)}}\\
	
	\cmidrule(lr){3-6} \cmidrule(lr){7-10}
	
	& &
	\multicolumn{2}{c}{\large\textbf{Direct}}&
	\multicolumn{2}{c}{\large\textbf{Inverse}}&
	\multicolumn{2}{c}{\large\textbf{Direct}}&
	\multicolumn{2}{c}{\large\textbf{Inverse}}\\ 
	
	\cmidrule(lr){3-4} \cmidrule(lr){5-6} \cmidrule(lr){7-8} \cmidrule(lr){9-10}
	
	\textbf{Fold \#} & \textbf{Test P\#}
	& \textbf{FLD} & \textbf{BLD} & \textbf{FLD} & \textbf{BLD}
	& \textbf{FLD} & \textbf{BLD} & \textbf{FLD} & \textbf{BLD} \\
	
	\midrule
	
	\textbf{1}  & 090,275 & 0.2439 & 0.2367 & 0.2775 & 0.2401 & 0.5000 & 0.3153 & 0.3708 & 0.5747 \\
	\textbf{2}  & 091,207 & 0.2406 & 0.2174 & 0.2692 & 0.2397 & 0.6429 & 0.2690 & 0.2512 & 0.6902 \\
	\textbf{3}  & 139,143 & 0.2232 & 0.2171 & 0.2704 & 0.2259 & 0.3091 & 0.2687 & 0.2710 & 0.2638 \\
	\textbf{4}  & 093,187 & 0.1984 & 0.2125 & 0.2641 & 0.2252 & 0.3485 & 0.2458 & 0.2400 & 0.4382 \\
	\textbf{5}  & 272,277 & 0.2390 & 0.2418 & 0.2702 & 0.2312 & 0.3119 & 0.2281 & 0.2856 & 0.3770 \\
	\textbf{6}  & 092,201 & 0.2351 & 0.2210 & 0.2748 & 0.2291 & 0.2236 & 0.2350 & 0.2429 & 0.2745 \\
	\textbf{7}  & 144,278 & 0.2307 & 0.2166 & 0.2675 & 0.2360 & 0.3484 & 0.2328 & 0.2950 & 0.3608 \\
	\textbf{8}  & 094,188 & 0.2252 & 0.2228 & 0.2639 & 0.2267 & 0.6567 & 0.3454 & 0.3602 & 0.4967 \\
	\textbf{9}  & 142,145 & 0.2120 & 0.2046 & 0.2525 & 0.2225 & 0.2901 & 0.2867 & 0.3414 & 0.2676 \\
	\textbf{10} & 141,205 & 0.2205 & 0.2170 & 0.2628 & 0.2260 & 0.5114 & 0.4143 & 0.3124 & 0.3639 \\
	
	\midrule
	
	\textbf{Avg} & & 0.2269 & 0.2208 & 0.2673 & 0.2302 & 0.4143 & 0.2841 & 0.2971 & 0.4107 \\
	\textbf{Std} & & 0.0134 & 0.0104 & 0.0067 & 0.0060 &  0.1448 & 0.0564 & 0.0455 & 0.1340 \\
	
	\bottomrule
	
\end{tabular}
}
\end{table}

To save computational resources, all numerical experiments described so far were conducted in a ``fixed'' scenario, namely for the same random initialization of the trainable parameters and reserving the same patients (\emph{P\#093}, \emph{P\#278}) to testing. This way of proceeding prevents from thoroughly assessing robustness, which is instead of paramount importance in DL applications. To this aim, we perform a 10--fold cross--validation, designed with respect to the ``original'' geometries in the dataset. This means that, if the anatomy of a given patient is reserved for testing, then all the augmented versions of such anatomy are not considered for training. For each fold, we run three independent training processes, considering different random seeds. For this test, both training and testing errors are computed solely accounting for original anatomies. \\

Table~\ref{tab: results CV} reports the obtained results, in terms of training and testing FLD and BLD, for both the inverse and the direct mapping. We observe that all models yield precise approximations of the diffeomorphic maps on the training datapoints, with maximal pointwise errors that always lie below the $0.30 \ \si{\centi\metre}$ threshold. However, markedly larger errors are produced at testing, in particular for folds \#1, \#2, \#8, \#10. This phenomenon can be explained by considering
that these folds respectively reserve for testing patients \emph{P\#275}, \emph{P\#207}, \emph{P\#188},
\emph{P\#205}, whose geometries present features that are uniquely represented within the dataset.
For instance (see Figure~\ref{fig: dataset}): patient \emph{P\#207} is characterized by the only anatomy whose RSA bends towards (and
not away from) RCCA; patient \emph{P\#205} is the only one whose horizontal LSA chunk could
not be segmented. Hence, the drop in precision can be ascribed to data paucity.

\subsection*{Test 5: Latent space analysis and generative modelling} 

\begin{figure}[!t]
\centering
\includegraphics[width=0.88\textwidth]{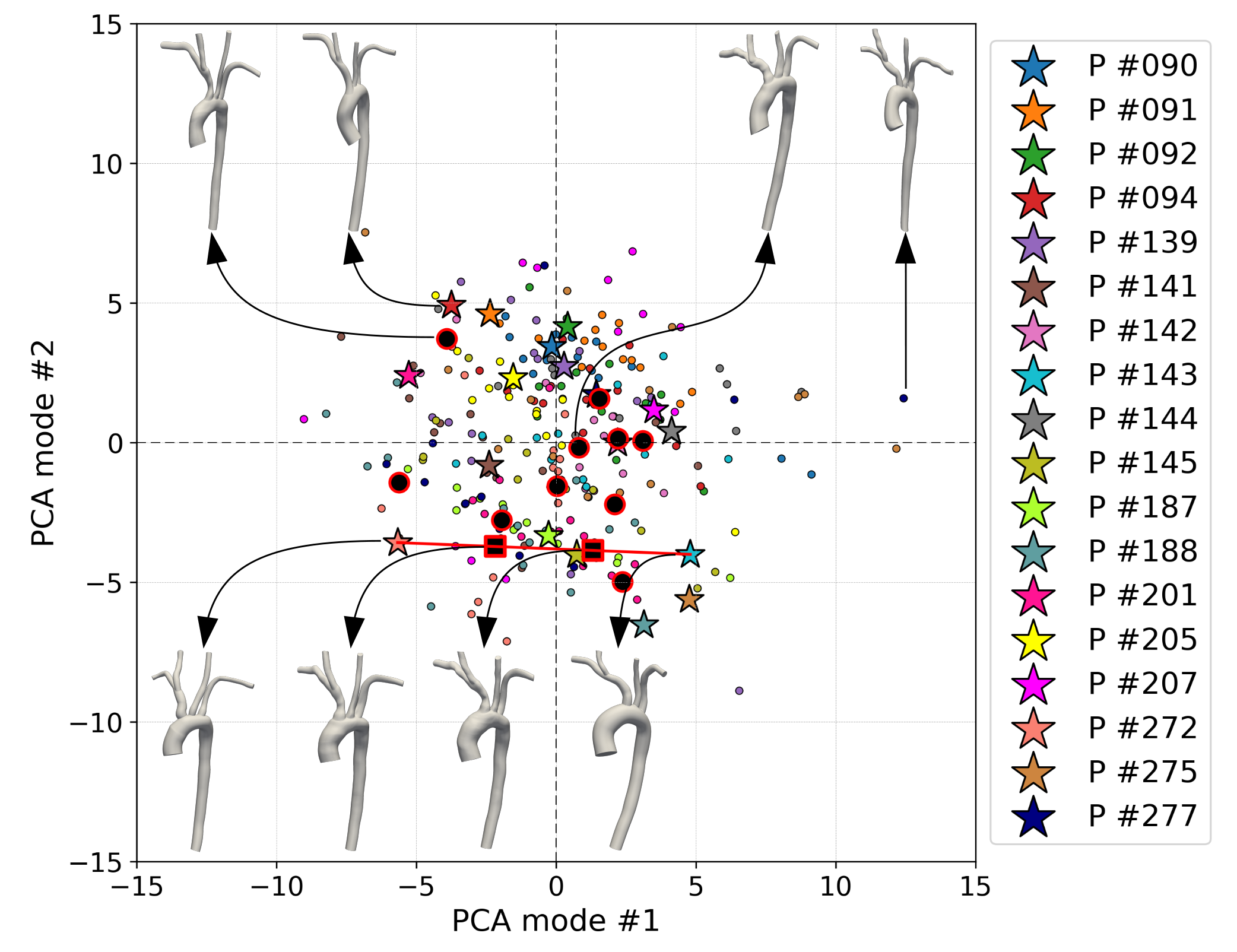} 
\caption{Representation of the latent space learned by the AD--SVFD model. In particular, we show the projection of the shape codes onto the two--dimensional subspace obtained through PCA on the whole set of training codes. We report the latent codes of the original patients (stars), and, for each of those, the latent codes of $10$ associated augmented geometries (circles). Furthermore, we display $10$ entries sampled from $\mathcal{N}(\bm{0}, \Sigma_z)$, where $\Sigma_z$ is an unbiased estimate of the covariance matrix computed from the training codes (red\&black circles), and $2$ entries sampled by linear interpolation between two patients in the latent space (red\&black squares). The black arrows map shape code instances to their counterparts in the physical space.}
\label{fig: latent codes PCA}
\end{figure}

The use of low--dimensional latent codes, belonging to the learnable space $\mathcal{Z}$, makes AD--SVFD suited for generative modelling. Indeed, once the model is trained, new anatomies can be generated by sampling shape code instances from $\mathcal{Z}$ and applying the corresponding inverse maps to the template geometry. Incidentally, we highlight that the robustness of the generative process is intimately related to the latent space regularity. For this reason, we include a penalization of the shape code entries in the loss, weighted by the positive constant $\omega_z$ (see Eq.~\eqref{eq: regularization term}).\\

Figure~\ref{fig: latent codes PCA} reports a sketch of the latent space learned by the AD--SVFD model. We show the projections of the shape codes onto a two--dimensional subspace, obtained through Principal Component Analysis (PCA). Furthermore, we display $10$ entries that are randomly sampled from $\mathcal{N}(\bm{0}, \Sigma_z)$ --- where $\Sigma_z$ is an unbiased estimate of the covariance matrix computed from the training shape codes (red\&black circles) --- and $2$ entries sampled by linear interpolation in the latent space (red\&black squares). From a qualitative standpoint, the learned space seems rather smooth. Indeed, geometries whose codes are close in $\mathcal{Z}$ also look similar in the physical space, whereas shapes whose codes lie far apart in the latent space exhibit evident discrepancies. The linear interpolation results (see bottom--left corner in Figure~\ref{fig: latent codes PCA}) support this observation, since the two sampled geometries feature intermediate traits between those of the two patients.


\section*{Discussion}  \label{sec: discussion}

We introduced AD--SVFD, a deep learning model for the non--rigid registration and synthetic generation of three--dimensional surfaces, tailored to vascular anatomies and, in particular, to healthy aortas. \\

Analogously to \cite{croquet2021unsupervised, amor2022resnet}, the AD--SVFD model performs 3D point cloud registration, leveraging shape representations in the form of weighted point clouds, whose weights are proportional to the nearest neighbours distance (see Eq.\eqref{eq: point clouds}). In this regard, AD--SVFD differs from deformable registration models based on continuous signed distance functions (SDFs), such as the ones presented in \cite{sun2022topology, kong2024sdf4chd}. As empirically demonstrated in \emph{Test 3} through a comparison of the performances of AD--SVFD and SDF4CHD, this approach enables more precise reconstructions, at least for vascular anatomies. Indeed, as shown in Figure~\ref{fig: comparison results}, AD--SVFD clearly outperforms SDF4CHD~\cite{kong2024sdf4chd}, whose deformed anatomies either omit or severely distort most of the smallest branches. This phenomenon can plausibly be attributed to the use of SDFs, whose resolution must remain limited for computational efficiency reasons, thereby hindering the accurate capture of the finest details. Notably, the outcomes of \emph{Test 3} also reveal two additional key aspects. On the one hand, AD--SVFD demonstrates superior performance, in both accuracy and efficiency, compared to alternative 3D point cloud registration methods such as ResNet--LDDMM~\cite{amor2022resnet}. On the other hand, traditional approaches like LDDMM~\cite{bone2018deformetrica}, not rooted in DL techniques, exhibit comparable accuracy metrics, but are significantly more computationally demanding at inference. \\

Dealing with point clouds in the absence of ground--truth point--to--point correspondences required the consideration of alternative data attachment measures, both to construct an effective loss function and to design informative error indicators. This aspect was analysed in \emph{Test 2}, where multiple data adherence metrics were investigated. Although representing the baseline alternative, the canonical (i.e. unweighted) Chamfer Distance outperforms all other options, delivering the most precise geometry reconstructions at the lowest computational costs and memory requirements. In the test case at hand, neither incorporating the normals' orientation nor exploiting the point cloud weights resulted in improved precision, while instead inducing moderate to substantial increases in complexity. Notably, the performance achieved using the debiased Sinkhorn divergence in the loss proved unsatisfactory in terms of both accuracy and efficiency, as also observed in \cite{amor2022resnet} on non--elementary geometries.\\

As in the model proposed in \cite{croquet2021unsupervised}, AD--SVFD expresses diffeomorphic maps through the \emph{stationary vector field} parametrization. Specifically, the ambient space deformation is defined as the solution at unit time of a system of ODEs, whose learnable right--hand side does not explicitly depend on time (see Eq.\eqref{eq: SVF ODE}). In particular, the right--hand side is modeled by a fully--connected and \emph{Leaky--ReLU} activated ANN, so to ensure well--posedness. By numerically integrating the diffeomorphic flow equations over time, it becomes possible to reconstruct the geodesic paths connecting source anatomies to the template. As illustrated in Figure~\ref{fig: zs error trends} (bottom), this procedure gives rise to a collection of synthetic shapes, exhibiting a smooth and gradual transition from the source characteristics to the reference ones. \\

Extracting the intermediate stages of numerical integration is not the only means of generating artificial geometries with AD--SVFD. Indeed, a crucial feature of the proposed model, distinguishing it for instance from ResNet--LDDMM~\cite{amor2022resnet}, is the internalization of latent embeddings for the source anatomies. Similarly to the models presented in \cite{sun2022topology, kong2024sdf4chd}, this is accomplished by introducing low--dimensional shape codes, which serve as trainable input variables within an auto--decoder architecture. Consequently, the available geometries are somehow non--linearly projected onto a low--dimensional latent space, where convenient random sampling routines can be implemented for generative purposes. Further details regarding the definition and treatment of shape codes are provided in \emph{Methods}. In \emph{Test 1}, it was demonstrated that the dimension of the latent space, denoted as $N_z$, should be carefully calibrated to optimally balance accuracy and efficiency. As illustrated in Figure~\ref{fig: zs error trends} (top), accuracy is significantly compromised when excessively small shape codes are employed, while it plateaus for large values of $N_z$, where model complexity and memory demands become instead prohibitive. \\

In addition to controlling the latent space dimension, monitoring its regularity is of paramount importance to ensure the robustness and reliability of downstream generative AI applications. To this end, a suitable penalization term was included in the loss function (see Eq.\eqref{eq: loss function}), with its weight $w_z=10^{-3}$ carefully fine--tuned. As briefly discussed in \emph{Test 5}, and illustrated in Figure~\ref{fig: latent codes PCA}, this strategy ultimately enables the construction of a smooth latent space that can be robustly queried to generate customizable, realistic, synthetic anatomies. It is worth noting that a well--established and widely adopted technique to enforce latent space regularity consists of variational training. Accordingly, several exploratory experiments were conducted in this direction, updating both the model structure and the loss function to implement a variational auto--decoder formulation for AD--SVFD~\cite{zadeh2019variational}. However, no substantial improvements in regularity or robustness were observed, while approximation quality was markedly degraded. \\

Despite exhibiting highly promising results, the current work nonetheless presents certain limitations. First and foremost, as with many DL--based models, data availability imposes non--negligible performance constraints, which could only be partially mitigated through data augmentation. This issue becomes particularly evident from the cross--validation results reported in \emph{Test 4}. Specifically, AD--SVFD accuracy on unseen anatomies declines for folds containing testing shapes featuring unique traits within the dataset, such as \emph{P\#205} and \emph{P\#207} (see Figure~\ref{fig: dataset}). It is noteworthy that additional aortic anatomies from the \emph{Vascular Model Repository} were also considered during preliminary stages. However, these were subsequently discarded due to incompatibility with the adopted data augmentation pipeline, which produced undesired non--physiological artifacts.
Incidentally, although the conducted tests were limited to healthy aortas, it is important to emphasize that the proposed registration approach is general and can be seamlessly extended to a wide range of challenging applications.
Secondly, to reduce computational effort, most hyperparameters were fine--tuned within a simplified single shape--to--shape registration setting, as detailed in \emph{Appendix}~\ref{app: TPE tuning}. In practice, only the hyperparameters associated with the shape codes ($N_z$, $w_z$, and $\lambda_z$) were calibrated using the full AD--SVFD model. Consequently, at least marginal performance improvements may be achievable through hyperparameters configurations specifically tailored to a multi--shape context. 
Lastly, the present analysis focused on the size and regularity of the latent space, but it did not address its interpretability. Investigating this aspect may substantially enhance the generative pipeline, and will therefore be the subject of future developments. \\

In conclusion, AD--SVFD may serve as a valuable tool for engineering applications involving physical problems in complex geometries. The proposed approach introduces potentially distinctive elements for facilitating geometry manipulation, notably by simultaneously providing compact and portable representations and by learning accurate, smooth, invertible, and topology--preserving mappings to a pre--defined reference. Notwithstanding improvements of its generalization capabilities, AD--SVFD is envisioned as a pre--trained module within physics--aware machine learning frameworks, enabling the incorporation of realistic geometrical variability into physical processes simulations \cite{sun2020surrogate, oldenburg2022geometry, regazzoni2022universal, costabal2024delta, brivio2025handling}.  


\section*{Methods}  \label{sec: methods}
We provide a more detailed analysis of the AD--SVFD model, specifically focusing on the shape codes, the ANN architecture, the numerical integration of the flow equations, the data attachment measures and the optimization procedure. 

\subsubsection*{Latent shape codes}
AD--SVFD provides a unified framework for the simultaneous registration of the source anatomies to a pre--defined template, leveraging implicit neural representations. Indeed, every source shape $\mathcal{S}_i$ is associated to a shape code $\bm{z}_i \in \R^{N_z}$, so that the diffeomorphism $\vec{\varphi}_i$ mapping $\mathcal{S}_i$ to $\mathcal{T}$ configures as the specialized version of a ``generic'' diffeomorphism $\vec{\varphi}$, i.e. $\vec{\varphi}_i(\bm{x}) := \vec{\varphi}(\bm{x}, \bm{z}_i)$, with $\bm{x} \in \R^3$. \\

Instead of directly providing the latent codes in input to the model, we borrow from \cite{kong2024sdf4chd} the use of a position--aware shape encoding strategy~\cite{chen2022unist}. Given a shape code $\bm{z}_i$, we define the associated shape code grid $\bm{Z}_i \in \R^{g_z \times g_z \times g_z \times (N_z/g_z^3)}$ as $\bm{Z}_i = \mathcal{R}_z(\bm{z}_i)$, where $\mathcal{R}_z: \R^{N_z} \to \R^{g_z \times g_z \times g_z \times (N_z/g_z^3)}$ is a suitable reshaping function. Here we suppose that the shape code dimension $N_z$ is a multiple of $g_z^3$; in this work, we always select $g_z=2$. Then, for a given point $\bm{x} \in \Omega$, the position--aware shape code $\bar{\bm{z}}_i(\bm{x}) \in \R^{N_z/g_z^3}$, associated to the source shape $\mathcal{S}_i$, is obtained by evaluating the trilinear interpolation of $\bm{Z}_i$ at $\bm{x}$, i.e. $\bar{\bm{z}}_i(\bm{x}) := \operatorname{Lerp}(\bm{x}, \bm{Z}_i)$, being $\operatorname{Lerp}(\cdot, \cdot)$ the trilinear interpolation function. This approach comes with two major advantages. On the one hand, the positional--awareness of the latent codes helps the model in better differentiating the deformation flow field, depending on the location within the domain. On the other hand, even if the total number of trainable parameters is unchanged, only $(N_z/g_z^3)$--dimensional vectors are provided as input to the ANN. Hence, model complexity is (slightly) reduced compared to the \emph{naive} approach, ideally at no loss in representation power.

\subsubsection*{Artificial neural network architecture}
To learn the diffeomorphisms between the source anatomies and the template, we exploit the Neural ODE approach~\cite{chen2018neural}, employing an ANN to approximate the right--hand side of Eq.\eqref{eq: SVF ODE}. More specifically, we consider a DL--based structure comprising three modules:
\begin{itemize}
\item \emph{Feature Augmentation network (FA--NN)}: the first part of the model performs a data--driven feature augmentation of the input locations. It consists of a fully--connected ANN that takes as input a spatial location $\bm{x} \in \Omega$ and the associated position--aware shape code $\bar{\bm{z}}_i(\bm{x})$ and yields a set of latent features $\bm{x}_{i, FA} \in \R^{N_{FA}}$ as output. Since FA--NN solely acts as a feature augmentation compartment, we do not want it to weigh down model complexity. So, we consider shallow networks with few neurons per layer. We highlight that the learned features are anatomy--dependent, thanks to the conditioning effect of the shape code on the model weights. We can summarize the FA--NN action via the function $\mathcal{F}_{FA}: \R^3 \times \R^{N_z/g_z^3} \to \R^{N_{FA}}$, such that $\bm{x}_{i, FA} = \mathcal{F}_{FA}(\bm{x}, \bar{\bm{z}}_i(\bm{x}); \bm{\Theta})$.
\item \emph{Fourier Positional Encoder (FPE)}: in the second module, the learned latent features $\bm{x}_{i, FA}$ undergo a further augmentation step, via a deterministic Fourier positional encoding~\cite{tancik2020fourier}, adopting a base--2 logarithmic sampling strategy in the frequency domain. This step is crucial for mitigating the spectral bias of ANNs~\cite{rahaman2019spectral}. We can summarize the FPE action via the function $\mathcal{F}_{FPE}: \R^{N_{FA}} \to \R^{N_{NPE}}$, such that $\bm{x}_{i,FPE} = \mathcal{F}_{FPE}(\bm{x}_{i, FA})$, where $N_{NPE} := (2N_e + 1) N_{FA}$.
\item \emph{Diffeomorphic Flow network (DF--NN)}: the last chunk of the ANN model is responsible for the approximation of the stationary velocity field at the spatial location $\bm{x}$ --- the right--hand side of Eq.\eqref{eq: SVF ODE} --- given the augmented features $\bm{x}_{i,FPE}$ and the position--aware shape code $\bar{\bm{z}}_i(\bm{x})$. As for FA--NN, we use a fully--connected ANN. However, since DF--NN is the core part of the model, we consider deeper architectures with a larger number of neurons in each layer. We underline that the output of DF--NN depends on the source anatomy, thanks to the external conditioning effect of the shape codes. We can summarize the DF--NN action via the function $\mathcal{F}_{DF}: \R^{N_{FPE}} \times \R^{N_z/g_z^3} \to \R^3$, such that $\bm{v}_i = \mathcal{F}_{DF}(\bm{x}_{i, FPE}, \bar{\bm{z}}_i(\bm{x}); \bm{\Theta})$.
\end{itemize} 

Ultimately, we can express the action of the entire ANN model by the function $\mathcal{F}: \R^3 \times \R^{N_z/g_z^3} \to \R^3$, defined as $\mathcal{F} := \mathcal{F}_{FA} \circ \mathcal{F}_{FPE} \circ \mathcal{F}_{DF}$.
As resulting from the hyperparameters tuning procedure (see \emph{Appendix}~\ref{app: HP tuning}), we consider: $(i)$~\emph{Leaky--ReLU} activation functions, with negative slope of $0.2$, for both FA--NN and DF--NN; $(ii)$~FA--NN with $3$ layers of dimension $64$; $(iii)$~DF--NN with $5$ layers of dimension $256$; $(iv)$~$N_e=3$ for the FPE. Neglecting the latent codes' contributions, the ANN model counts approximately $278k$ trainable parameters.
To ease the notation, in the following we omit the explicit spatial dependency of the shape codes. 

\subsubsection*{Numerical integration of the flow equations}
For the numerical integration of the diffeomorphic flow equations (see Eq.\eqref{eq: SVF ODE}), we rely on first--order methods. Specifically, for the forward--in--time direct mapping, we employ the explicit forward Euler method. Let $\bm{x}_{i,j}^{s, (0)} = \bm{x}_{i,j}^s \in \Omega$ be a point of the source point cloud $\mathcal{S}_i$, where the superscript $(0)$ denotes the initial iteration count. Also, let $K \in \N$ be the number of time steps; in all tests, we set $K=10$. Then, for $k < K$, the time marching scheme proceeds as follows:
\begin{equation}
\label{eq: direct mapping forward euler}
\bm{x}_{i,j}^{s, (k+1)} = \bm{x}_{i,j}^{s, (k)} + \frac{1}{K} \mathcal{F}\left(\bm{x}_{i,j}^{s, (k)}, \bar{\bm{z}}_i; \bm{\Theta}\right) = \bm{x}_{i,j}^{s, (k)} + \frac{1}{K} \bm{v}_{i,j}^{s, (k)}~.
\end{equation}
The point corresponding to $\bm{x}_{i,j}^s$ in the template space is then the result of Eq.\eqref{eq: direct mapping forward euler} at $k=K-1$, i.e $\bm{x}_{i,j}^{s, (K)}$. \\

To compute the inverse map, which deforms the ambient space so as to overlap the template anatomy to the source, we integrate the flow equations backward--in--time, given a final condition. In particular, we want the discrete inverse map to be the ``true'' inverse of the discrete direct map, defined in Eq.\eqref{eq: direct mapping forward euler}. So, let $\bm{x}_{i,j}^{t, (K)} = \bm{x}_{j}^t \in \Omega$ be a point of the template point cloud $\mathcal{T}$. The time marching scheme at step $k > 0$ proceeds as follows:
\begin{equation}
\label{eq: inverse mapping backward euler}
\bm{x}_{i,j}^{t, (k-1)} = \bm{x}_{i,j}^{t, (k)} - \frac{1}{K} \mathcal{F}\left(\bm{x}_{i,j}^{t, (k-1)}, \bar{\bm{z}}_i; \bm{\Theta}\right)~.
\end{equation}
Even though Eq.\eqref{eq: inverse mapping backward euler} allows to invert Eq.\eqref{eq: direct mapping forward euler} exactly, its use may be difficult in practice, being an implicit scheme. Indeed, the nonlinearity of $\mathcal{F}$ entails the use of \emph{ad hoc} numerical techniques, such as Newton iterations, to compute a solution. Despite the Jacobian of $\mathcal{F}$ can be efficiently computed by automatic differentiation, the whole procedure is likely to slow down both the forward and the backward pass. For this reason, we rely on a first--order explicit approximation of Eq.\eqref{eq: inverse mapping backward euler} --- known as the modified Euler scheme --- that writes as follows:
\begin{equation}
\label{eq: inverse mapping modified euler}
\bm{x}_{i,j}^{t, (k-1)} = \bm{x}_{i,j}^{t, (k)} - \frac{1}{K} \mathcal{F}\left(\bm{x}_{i,j}^{t, (k)} - \frac{1}{K} \mathcal{F}\left(\bm{x}_{i,j}^{t, (k)}, \bar{\bm{z}}_i; \bm{\Theta}\right), \bar{\bm{z}}_i; \bm{\Theta}\right) =  \bm{x}_{i,j}^{t, (k)} - \frac{1}{K} \bm{v}_{i,j}^{t, (k)}~.
\end{equation}
The point corresponding to $\bm{x}_j^t$ in the source space is then the result of Eq.\eqref{eq: inverse mapping modified euler} at $k=1$, i.e $\bm{x}_{i,j}^{t, (0)}$. 

\subsubsection*{Data attachment measures} 
As reported in Eq.\eqref{eq: point clouds}, we represent three--dimensional surfaces as (weighted) point clouds and we assume to not know exact point--to--point correspondences. Therefore, suitable data attachment measures to quantify the discrepancy between point clouds have to be considered. The simplest alternative is offered by the Chamfer Distance (CD) $\mathcal{D}_{CD}: \R^{M \times 3} \times \R^{M' \times 3} \to \R^+$, that is defined as 
\begin{equation}
\label{eq: chamfer distance}
\mathcal{D}_{CD}(Y, Y') := 
\frac{1}{M} \sum_{i=1}^{M}\min_{\bm{c'} \in Y'} \norm{ Y_i - \bm{c'} }_2^2 \ + \ 
\frac{1}{M'} \sum_{i'=1}^{M'} \min_{\bm{c} \in Y} \norm{ \bm{c} - Y'_{i'} }_2^2~.
\end{equation}
In particular, the CD comprises the sum of two terms: the forward CD (FCD), which compares the points in $Y$ with the closest ones in $Y'$, and the backward CD (BCD), which compares the points $Y'$ with the closest ones in $Y$. Considering both components is crucial to obtain a meaningful goodness--of--fit measure. CD has proven to be an effective metric for diffeomorphic registration, particularly in the computational anatomy framework, as shown e.g. in \cite{amor2022resnet}. However, in \cite{feydy2017optimal} it has been demonstrated that using CD is also likely to yield low--quality gradients. To mitigate this issue, we consider the Earth's Mover Distance (EMD) $\mathcal{D}_{EMD}: \R^{M \times 3} \times \R^{M' \times 3} \to \R^+$ \cite{rubner2000earth, sinha2024deepemd}:
\begin{equation*}
\label{eq: earth mover distance}
\mathcal{L}_{EMD} (Y, Y') = \min_{\xi \in \mathcal{M}(Y, Y')} \sum_{\bm{y} \in Y} \vert \vert \bm{y} - \xi(\bm{y}) \vert\vert_2^2~,
\end{equation*}
where $\mathcal{M}(Y, Y')$ denotes the set of 1--to--1 (bipartite) mappings from Y to Y'.
In a nutshell, EMD is a Wasserstein distance, that seeks for the optimal transport plan that orders the points in $Y'$ to match the ones in $Y$. In practice, we approximate EMD with the debiased Sinkhorn Divergence (SD) $\mathcal{D}_{SD}$~\cite{cuturi2013sinkhorn}. The latter is the solution to an Optimal Transport problem with entropic constraints, and it can be estimated using the iterative Sinkhorn's algorithm \cite{chizat2020faster}. We refer the reader to \cite{feydy2017optimal} for the precise definition of $\mathcal{D}_{SD}$; further details and a state--of--the--art literature review on diffeomorphic registration using SD can be found in \cite{delara2023diffeomorphic}. In all numerical tests conducted using SD, we consider a quadratic ground cost point function, a temperature scalar $\varepsilon = 10^{-4}$, and a linear $\varepsilon$--scaling with factor $0.9$. This combination of hyperparameters should be sensible for input measures that lie in the unit cube, providing a good trade--off between accuracy and efficiency~\cite{feydy2019interpolating}. \\

According to \cite{feydy2017optimal}, SD is a good overlapping metric only if the points are roughly equispaced. However, SD can be effectively extended to unevenly distributed point clouds if the latter are weighted, i.e. if each point is associated to a quantity proportional to its distances from the closest neighbours. In fact, such weights appear in the entropic regularization term and in the entropic constraints of the associated optimal transport problem, awarding more ``importance'' to the most isolated points in the cloud. As reported in \emph{Introduction}, in this work we extract the cloud points as the cell centers of available surface triangulations and we compute the weights as the corresponding cell areas, normalized to add up to one. In the following, we denote by $\mathcal{D}_{SD}$ the standard SD, where all the weights are assumed to be equal, and by $\mathcal{D}^W_{SD}$ the weighted SD. A similar reasoning can also be extended to the CD, even if the latter is not related to any optimal transport problem~\cite{wu2021density}. In this work, we define a weighted CD $\mathcal{D}^W_{CD}: \R^{M \times (3+1)} \times \R^{M' \times (3+1)} \to \R^+$ as follows:
\begin{equation}
\label{eq: weighted chamfer distance}
\mathcal{D}^W_{CD}((Y,w), (Y',w')) := \frac{1}{N} \sum_{i=1}^{M} w_i \min_{\bm{c'} \in Y'} \norm{ Y_i - \bm{c'} }_2^2 \ + \ \frac{1}{N'} \sum_{i'=1}^{M'} w'_i \min_{\bm{c} \in Y} \norm{ \bm{c} - Y'_{i'} }_2^2~.
\end{equation}

Alternatively to the use of SD, we also try to mitigate the low--quality gradient issue by developing variants of CD that exploit information coming from the source and template surface normals. In fact, CD is agnostic of the closed surface structure of the manifold from which the points are sampled, as it solely relies on point--to--point distances. In particular, we consider two surface--aware corrections of CD. 
The first one --- denoted as $\mathcal{D}_{NCD}$ --- simply consists of adding a regularization term that penalizes the discrepancy between the normals, i.e.
\begin{equation}
\label{eq: normals chamfer distance}
\mathcal{D}_{NCD}(Y, Y') := \mathcal{D}_{CD}(Y,Y') \ + \  \frac{w_n}{2M} \sum_{i=1}^{M} \Big(1 - \bm{n_i} \cdot \bm{n_{c'_i}}\Big)^2 \ + \ 
\frac{w_n}{2M'} \sum_{i'=1}^{M'} \Big(1 - \bm{n_{i'}} \cdot \bm{n_{c_{i'}}}\Big)^2~,
\end{equation}
where $\bm{c'}_i := \min_{\bm{c'} \in Y'} \norm{ Y_i - \bm{c'} }_2^2 $, $\bm{c}_{i'} := \min_{\bm{c} \in Y} \norm{ \bm{c} - Y'_{i'} }_2^2 $, and $w_n \in \R^+$ is a scale factor. Here $\bm{n_i}, \bm{n_{i'}}, \bm{n_{c'_i}}, \bm{n_{c_{i'}}} \in \R^3$ denote the (supposed known) outward unit normal vectors to the target surface, evaluated at $Y_i, Y'_i, \bm{c'}_i, \bm{c}_{i'}$, respectively. Also, $\bm{a}\cdot\bm{b} := \sum_j \bm{a}_j \bm{b}_j$ is the standard inner product. Preliminary numerical tests suggested to set $w_n = 10^{-2}$, which results in the normals' penalization term to account for roughly $10 \%$ of the loss value.
The second variant of CD, instead, is offered by the point--to--plane CD (PCD)~\cite{tian2017geometric}, denoted as $\mathcal{D}_{PCD}$ and defined as
\begin{equation*}
\label{eq: point to plane chamfer distance}
\mathcal{D}_{PCD}(Y, Y') := \frac{1}{N} \sum_{i=1}^{M} \min_{\bm{c'} \in Y'} \left(\left(Y_i - \bm{c'}\right) \cdot \bm{n_i} \right)^2 + \frac{1}{N'} \sum_{i'=1}^{M'} \min_{\bm{c} \in Y} \left(\left(\bm{c} - Y'_i\right) \cdot \bm{n_{i'}} \right)^2 ~,
\end{equation*}
where $\bm{n_i}, \bm{n'_i}$ are as in Eq.\eqref{eq: normals chamfer distance}. The PCD computes the error projections along the normal directions, thus solely penalizing points that ``move away'' from the target local plane surface. For point clouds that are sampled from surfaces, this distance is better aligned with the perceived overlapping quality than the canonical CD. 
Analogously to $\mathcal{D}^W_{CD}$ defined in Eq.\eqref{eq: weighted chamfer distance}, weighted versions of $\mathcal{D}_{NCD}$ and $\mathcal{D}_{PCD}$, respectively denoted as $\mathcal{D}^W_{NCD}$ and $\mathcal{D}^W_{PCD}$, can be constructed. 

\subsubsection*{Training procedure}

\begin{algorithm}[t!]
	\caption{AD--SVFD model training pipeline}
	\label{alg: training algorithm}
	\begin{algorithmic}[1]
		\Procedure{Train\_AD\_SVFD}{$\mathcal{S}_1, \cdots, \mathcal{S}_{N^s}, \mathcal{T}, E, B, M$}   
		\Statex \Comment{\parbox[t]{.95\linewidth}{$\mathcal{S}_i$: $i$--th source point cloud; $\mathcal{T}$: template point cloud;  $E$: \# epochs; $B$: batch size; $M$: \# sampled points} }
		
		\bigskip    
		
		\State Initialize ANN parameters $\bm{\Theta}$
		\ForAll{$i \in \{i_1, \cdots, i_{N^s}\}$}
		\State $\mathcal{L}_i^s, \ \mathcal{L}_i^t \gets \bm{0}_{M}, \ \bm{0}_{M}$  \Comment{Initialize pointwise loss functions}
		\State Sample $\bm{z}_i^s \sim \mathcal{N}\left(\bm{0}, \frac{2}{N_z}I\right)$  \Comment{Initialize shape code}
		\EndFor
		\State $\mathcal{L}^t \gets \bm{0}_{M}$  
		
		\medskip
		
		\State $e \gets 0$
		\While{$e < E$}  \Comment{Loop over epochs}
		
		\State $b, \ \mathcal{B} \gets 0, \ [ \ ]$
		
		\medskip
		
		\While{$b < \left\lceil\frac{N^s}{B}\right\rceil$}  \Comment{Loop over batches}
		\State $\bar{B} \gets B+1 \ \ \text{if} \ \ b < (N^s \mathbin{\%} B) \ \ \text{else} \ \ B$  \Comment{Define batch size}
		\State Sample $i_1, \cdots, i_{\bar{B}} \sim \mathcal{U}(\{1, \cdots, N^s\}\setminus \mathcal{B})$ 
		
		\medskip
		
		\State $\mathcal{T}^b \gets \operatorname{PointSample}(\mathcal{T}, \mathcal{L}^t, M)$  \Comment{Sample template points}
		\ForAll{$i \in \{i_1, \cdots, i_{\bar{B}}\}$}
		\State $\mathcal{S}_i^b \gets \operatorname{PointSample}(\mathcal{S}_i, \mathcal{L}_i^s, M)$  \Comment{Sample source points}
		\State $\bar{\bm{z}}_i \gets \operatorname{CodeSample}(\bm{z}_i, \mathcal{S}_i^b)$  \Comment{Sample shape code}
		\State $\mathcal{S}_i^{b, (K)} \gets \mathcal{D}_{SVF}(\mathcal{S}_i^b, \bar{\bm{z}}_i, \bm{\Theta})$ \Comment{Direct mapping}
		\State $\mathcal{T}_i^{b, (0)} \gets \mathcal{I}_{SVF}(\mathcal{T}^b, \bar{\bm{z}}_i, \bm{\Theta})$  \Comment{Inverse mapping}
		\State $\mathcal{L}_i^s \gets \mathcal{L}(\mathcal{S}_i^{b, (K)}, \mathcal{T})$  \Comment{Direct mapping loss}
		\State $\mathcal{L}_i^t \gets \mathcal{L}(\mathcal{T}_i^{b, (0)}, \mathcal{S}_i)$  \Comment{Inverse mapping loss}
		\EndFor
		
		\medskip
		
		\State $\mathcal{L}_{tot} \gets \frac{1}{\bar{B} M} \sum_{i,j=1}^{\bar{B}, M} (\mathcal{L}_{i,j}^s + \mathcal{L}_{i,j}^t) + \mathcal{L}_{reg}$  \Comment{Total loss}
		
		\medskip
		
		\State $\bm{\Theta} \gets \operatorname{Update\_ANN}(\bm{\Theta}, \mathcal{L}_{tot})$  \Comment{Update ANN parameters}
		\ForAll{$i \in \{i_1, \cdots, i_{\bar{B}}\}$}
		\State $\bm{z}_i \gets \operatorname{Update\_Codes}(\bm{z}_i, \mathcal{L}_i^s, \mathcal{L}_i^t)$  \Comment{Update shape codes}
		\EndFor 
		
		\medskip
		
		\State $b, \ \mathcal{B} \gets b+1, \ [ \mathcal{B}, i_1, \cdots, i_{\bar{B}} ]$                 
		
		\EndWhile
		
		\medskip
		
		\State $\mathcal{L}^t \gets \frac{1}{N^s} \sum_{i=1}^{N^s} \mathcal{L}_i^t$  \Comment{Average template loss}
		
		\State $e \gets e + 1$
		
		\EndWhile
		
		\EndProcedure
	\end{algorithmic}
\end{algorithm}

The training pipeline of AD--SVFD is reported in detail in Algorithm~\ref{alg: training algorithm}. Hereafter, we only discuss a few relevant aspects. Algorithm~\ref{alg: training algorithm} features a two--stage sampling procedure over the training epochs. Firstly, since we employ a batched stochastic optimization algorithm, we sample uniformly at random (without replacement) $B$--dimensional batches of source shapes with the associated shape codes (line 12). Then, for each of the selected point clouds, we sample $M$--dimensional sub--clouds (line 15); we also sample a $M$--dimensional sub--cloud for the template anatomy (line 13). In all tests, we set $B=8$ and $M=2'000$. The motivation behind point clouds resampling is two fold. On the one hand, it makes the training algorithm complexity independent of the level of refinement in the data, which is of paramount importance if the cardinality of the original clouds is large. Indeed, the complexity of all considered data attachment measures is quadratic in the number of points. On the other hand, resampling can be interpreted as a form of data augmentation and as such it allows improving robustness. Furthermore, we remark that the trilinear interpolation to compute the position--aware shape codes is repeated at every epoch (line 16), and it is also recurrently performed during time integration of the diffeomorphic flow ODE. \\

To further improve model performance, when using data attachment measures that allow for a pointwise evaluation (such as the ones based on CD), we implement a simple adaptive sampling procedure. This explains the presence of the template and source pointwise loss functions as input arguments to \emph{PointSample} in lines 13 and 15, respectively. Specifically, at each training epoch we sample $\lceil(1-a)M\rceil$ points uniformly at random, whereas the remaining $\lfloor aM \rfloor$ points are retained from the previous epoch, being the ones associated to the highest loss values. In this way, we oversample regions featuring larger mapping errors, tentatively driving the model towards homogeneously accurate predictions in space.
In all tests, we consider $a=0.15$, as resulting from the calibration procedure reported in \emph{Appendix}~\ref{app: TPE tuning}.\\

\begin{remark}
If the chosen data attachment measure does not allow for a pointwise evaluation, because it yields a cumulative discrepancy value, adaptive sampling cannot be performed. For instance, this is the case with SD. In Algorithm~\ref{alg: training algorithm}, the point sampling routines no longer depend on the loss at the previous epoch (lines 13,15), and no averaging over the points in the clouds is necessary to compute the total loss (line 21).
\end{remark}

The joint optimization of the ANN parameters $\bm{\Theta}$ (line 22) and of the latent shape codes $\{\bm{z}_i\}_{i=1}^{N^s}$ (line 24) is achieved by minimizing the loss function reported in Eq.\eqref{eq: loss function}. Notably, to limit the kinetic energy of the system that connects the source to the target, thus encouraging minimal deformations and reducing the risk of overfitting, we introduce the regularization term $\mathcal{L}_{\operatorname{reg}}$:
\begin{equation}
\label{eq: regularization term}
\mathcal{L}_{reg}(\bm{\Theta}) \ := \ 
\sum_{i=1}^{N^s}  \sum_{j=1}^M \sum_{k=0}^{K-1} \left( \lVert \bm{v}_{i,j}^{s, (k)}(\bm{\Theta}) \rVert_2^2 + \lVert \bm{v}_{i,j}^{t, (k+1)}(\bm{\Theta}) \rVert_2^2 \right)~,
\end{equation} 
where $\bm{v}_{i,j}^{s, (k)}$, $\bm{v}_{i,j}^{t, (k)}$ are defined as in Eqs.\eqref{eq: direct mapping forward euler},\eqref{eq: inverse mapping modified euler}, respectively. For both $\bm{\Theta}$ and the latent codes, we perform random initialization, drawing values from a Kaiming normal distribution~\cite{he2015delving} and we rely on the first--order Adam optimizer~\cite{kingma2014adam} for the update step.
Hyperparameter calibration tests suggested adopting the same learning rate $\lambda = 10^{-3}$ for all the trainable parameters. Finally, we run $E=500$ training epochs, which guarantees convergence of the optimization procedure. \\

\begin{remark}
During inference, a pipeline similar to Algorithm~\ref{alg: training algorithm}, but much cheaper, is performed. Indeed, the optimization problem to be solved is much smaller, since just the $N_z$ latent code entries associated with a single unseen shape have to be optimized. Remarkably, the low memory requirements enable the use of more advanced and memory--intensive optimizers, such as L--BFGS~\cite{bottou2018optimization}, to fine--tune Adam predictions, attaining superlinear convergence rates.
\end{remark}


\section*{Acknowledgements}
RT and SD were supported by the \emph{Swiss National Science Foundation}, grant No 200021\_197021 -- ``Data--driven approximation of hemodynamics by combined reduced order modeling and deep neural networks''.

LP, FK, and AM were supported by the \emph{U.S. National Science Foundation}, grant No. 2310909 -- ``Collaborative Research: Frameworks: A multi-fidelity computational framework for vascular mechanobiology in SimVascular'', and grant No. 1663671 -- ``SI2-SSI Collaborative Research: The SimCardio Open Source Multi-Physics Cardiac Modeling Package''. FK was also supported by the \emph{National Institutes of Health} (R01EB029362, R01LM013120, R38HL143615).

FR and SP were supported by the grant \emph{Dipartimento di Eccellenza 2023-2027 of Dipartimento di Matematica, Politecnico di Milano}, and by the project \emph{PRIN2022, MUR, Italy, 2023-2025, P2022N5ZNP}, ``SIDDMs: shape--informed data--driven models for parametrized PDEs, with application to computational cardiology'', funded by the European Union (Next Generation EU, Mission 4 Component 2). FR and SP are members of GNCS, ``Gruppo Nazionale per il Calcolo Scientifico'' (National Group for Scientific Computing) of INdAM (Istituto Nazionale di Alta Matematica).  

\section*{Data availability}
The dataset employed for the current study is publicly available at \url{ https://doi.org/10.5281/zenodo.15494901}. All patient--specific anatomies are publicly available on the Vascular Model Repository (\url{https://www.vascularmodel.com/dataset.html}).

\section*{Code availability}
The underlying code for this study is currently not available, but may be made provided to qualified researchers upon request to the corresponding author. Future publications of the software are being considered to support transparency and reproducibility.

\bibliographystyle{unsrtnat}
\bibliography{references}

\clearpage

\appendix

\section{Data augmentation procedure}
\label{app: data augmentation}
Since the available dataset of healthy aortic anatomies is too restrained for seep learning (DL) applications, we implemented a data augmentation algorithm based on radial basis functions interpolation. More specifically, we considered thin--plate splines (TPS), a special case of polyharmonic splines introduced in \cite{duchon1977splines}, that admits a natural radial basis function representation via the infinite--support kernel function $\kappa(x) = x^2 \log x$.

\subsection{Deformable registration by TPS interpolation}
\label{subsec: deformable registration by thin--plate spline interpolation}
Let us consider a pair of geometries $(G_{\alpha}, G_{\beta})$. Let us suppose to know $M \in \N$ exact point--to--point correspondences $\{(\bm{x}_{\alpha}^j, \bm{x}_{\beta}^j)\}_{j=1}^M$. Then, TPS interpolation finds a diffeomorphism that deforms $G_{\alpha}$ into $G_{\beta}$ by solving the following energy minimization problem:
\begin{equation}
\label{eq: TPS energy minimization}
\begin{alignedat}{2}
\vec{g}_\star &= \ \argmin_{\vec{g} \in \mathcal{G}} \sum_{j=1}^M \norm{ \bm{x}_\beta^j - \vec{g}(\bm{x}_\alpha^j)}_2^2 \ + \ w_H \norm{ H_g(\bm{x}_\alpha^j) }_F^2~, \\
&\qquad \text{where} \quad \mathcal{G} := \Big\{ \vec{g}: \R^3 \to \R^3: \ \vec{g}(\vec{x}) = \sum_{j=1}^M g_j \ \kappa\left(\norm{ \vec{x} - \bm{x}_\alpha^j  }_2 \right)\Big\}~.
\end{alignedat}
\end{equation}
Here, $H_g: \R^3 \to \R^{3 \times 3}$ denotes the Hessian of $g$, and $\norm{\cdot}_F: \R^3 \to \R^+$ is the Frobenius norm operator. The smoothing parameter $w_H \in \R^+$ allows to balance the goodness of fit with the regularity of the deformation. The most relevant limiting factor to the use of TPS interpolation is the availability of reliable point--to--point correspondences.\\

In this work, we identify corresponding points by exploiting the peculiar structure of the geometries at hand. Indeed, vascular anatomies consist of the intersection of several vessels, each one featuring a tube--like shape. In \emph{SimVascular}~\cite{updegrove2017simvascular}, vessels are conveniently modelled by their centerline, which is approximated by a trivariate cubic spline, and by a number of surface contours, planar closed lines that define the cross--sectional vessel lumen boundary at selected locations along the centerline. Even if \emph{SimVascular} allows to accurately describe surface contours using B--splines, we relied on a much simpler approximation, supposing the cross--sectional areas to be circular and centered at the centerline points. Furthermore, to derive more precise point--to--point correspondences, we partitioned some of the vessel into chunks, which are defined depending on the location of eventual branches. Indeed, as displayed in Figure~\ref{fig: dataset}~(b) in the manuscript, each anatomy in the dataset features $N_v=5$ vessels (aorta, LSA, RSA, RCCA, LCCA), but $N_p=7$ vessel portions (AA, DA, BA, LSA, RSA, LCCA, RCCA). Now, let $M_p \in \N$ be the total number of points sampled in each vessel portion, and let $M_c \in \N$ be the number of points sampled at each contour. In this work, we consider $M_p = 250$ and $M_c = 4$. For a given vessel portion $p$ of $G_{\alpha}$, the sampled points $\{\bm{x}_{\alpha, p}^j\}$ are structured as follows:
\begin{itemize}
	\item $M_p / (M_c + 1)$ points are uniformly distributed along the centerline;
	\item $(M_c M_p) / (M_c + 1)$ points are uniformly distributed along the (approximated) circular contours, corresponding to each sampled centerline point.
\end{itemize}
The final set of sampled points is then given by $X_{\alpha} = \bigcup_{\ell=1}^{N_p} \{\bm{x}_{\alpha,p_{\ell}}^j\}_{j=1}^{M_p}$.
The same sampling strategy is used to define the set of sampled points $X_{\beta}$ for $G_{\beta}$. 

\begin{remark*}
	For every point in a child branch, we compute the convex hull generated by its $1,000$ nearest neighbours in the parent vessel. If the point belongs to the convex hull, it means that it lies inside the parent vessel and so it is removed from the dataset. Additionally, also the points that lie outside of the convex hull by a distance smaller than $\tau D_p$ are discarded, where $D_p \in \R^+$ is the maximal distance between two points in the child branch and $\tau \in \R^+$ is a prescribed threshold. This helps in guaranteeing the well--posedness of the TPS interpolation problem. Ultimately, the total number of points sampled at a vessel portion is $\mytilde{M}_p \leq M_p$. If a point is removed from $X_{\alpha}$, the corresponding one is removed from $X_{\beta}$, and viceversa.
\end{remark*}

While the correspondences quality for the centerline points is often remarkable, the same consideration does not hold for the ones sampled on the lateral surface. Indeed, since the centerline is an open curve, the knowledge of the curvilinear coordinates alone is sufficient to derive solid correspondences. However, the surface contours are planar closed curves; this entails that reliably corresponding samples can be selected only upon convenient choices of two--dimensional reference frames. In fact, the selection of matching surface samples closely depends on the identification of topologically equivalent zero--degree angles in the cross--sectional planes. To this aim, we employ the Bishop frame of reference~\cite{bishop1975there, ebrahimi2021low}, a coordinates system for curves, which is defined by transporting a given reference frame (forward and/or backward) along the curve itself. Two peculiarities of the Bishop frame are noteworthy. Firstly, one of its vectors always coincides with the curve tangent. Secondly, the coordinates system exhibits a uniform zero twist along the curve. Therefore, if we are able to define equivalent reference frames for the cross--sectional planes located at two corresponding centerline points, then such frames can be robustly ``extended'' to the whole vessel. In this work, the equivalent reference frames have been derived using \emph{ad hoc} techniques, based on the relative positions of inlets and outlets. For instance, the vector that defines the zero--degree angle at the aorta's inlet contour is computed as the orthogonal projection of the vector that connects the aorta's centerline endpoints. \\

In order to guarantee the quality of point--to--point correspondences, an initial rigid alignment of the geometries is crucial. To this aim, we employ the Coherent Point Drift (CPD) algorithm, a point set registration method based on Gaussian Mixture Models~\cite{myronenko2010point}. Compared to the most popular Iterative Closest Point (ICP) algorithm~\cite{besl1992method} and to its most widely employed variants and alternatives (such as Levenberg--Marquardt ICP~\cite{fitzgibbon2003robust} or Robust Point Matching~\cite{gold1998new, chui2000new}), CPD proved to be more accurate and robust in presence of noise, outliers and missing points. Nonetheless, CPD is an iterative algorithm, and hence its accuracy is strictly linked to the choice of a good initial guess. For this reason, prior to the execution of CPD, we perform the following three--steps \emph{ad hoc} rigid registration and rescaling procedure, as reported at line~6 in Algorithm~\ref{alg: data augmentation}.\\ 

Let $\mathcal{S}_{\alpha}$, $\mathcal{S}_{\beta}$ be two point clouds, computed from the surface meshes $\mathcal{M}_{\alpha}$, $\mathcal{M}_{\beta}$. Furthermore, let us suppose to know the position of the aorta's inlet center and the normal vector to the aorta's outlet, for both geometries. Then, we proceed as follows:
\begin{enumerate}
	\item  \emph{Rescaling}: translate $\mathcal{S}_{\alpha}$, so that its barycenter coincides with the one of $\mathcal{S}_{\beta}$. Perform an isotropic rescaling of $\mathcal{S}_{\alpha}$, so that the maximal distance between points in $\mathcal{S}_{\alpha}$ equals the one in $\mathcal{S}_{\beta}$. We call the output $\mathcal{S}_{\alpha}^{(1)}$.
	\item \emph{Translation}: translate $\mathcal{S}_{\alpha}^{(1)}$, so that its aorta's inlet center coincides with the one of $\mathcal{S}_{\beta}$. We call the output  $\mathcal{S}_{\alpha}^{(2)}$.
	\item \emph{Rotation}: rotate $\mathcal{S}_{\alpha}^{(2)}$ at the aorta's inlet center around the aorta's outlet normal vector by the angle $\vartheta$ that minimizes the Chamfer Distance (CD) between $\mathcal{S}_{\beta}$ and $\mathcal{S}_{\alpha}^{(2)}$. We call the output $\bar{\mathcal{S}}_{\alpha}$, which serves as the initial guess for the CPD iterations.
\end{enumerate}

\begin{figure}[t!]
	\centering
	\includegraphics[width=0.80\textwidth]{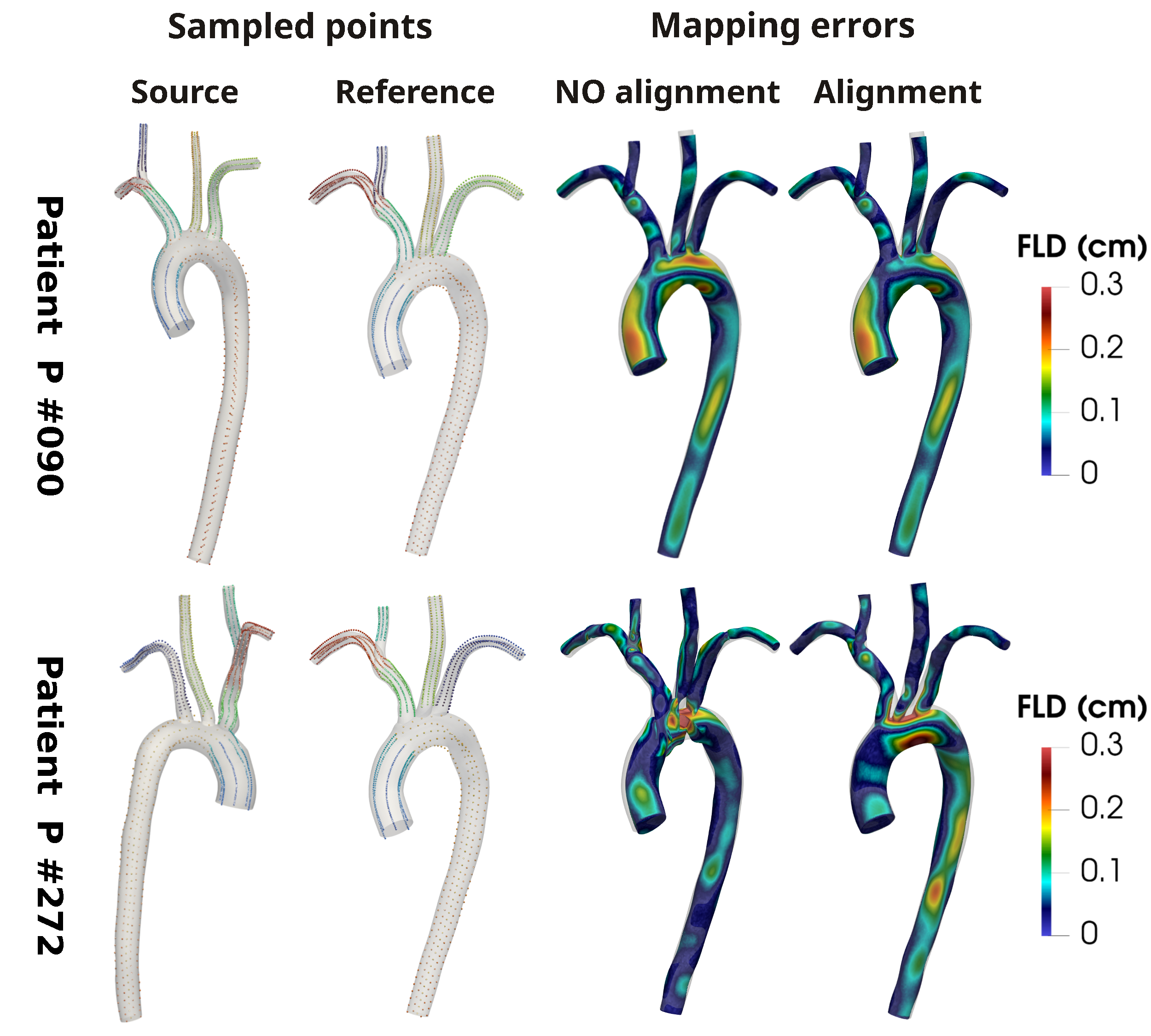}
	\caption{Visualization of the TPS interpolation results. In particular, for two patients in the dataset (P\#090 and P\#272) we show the locations of the interpolation points in the target and template geometries --- color--coded so that corresponding points share the same value --- and the pointwise mapping errors, quantified through the forward local distance (FLD), expressed in \si{\centi\metre}. For the mapping results, we compare the errors obtained without and with a preliminary rigid registration of the geometries to the template by the Coherent Point Drift algorithm. For reference, the template inlet diameter is $1.31$ \si{\centi\metre} for \emph{P\#091}.}
	\label{fig: RBF results}
\end{figure}

\begin{table}[t!]
	\centering
	\caption{TPS deformable registration errors. In particular, we report the average and maximal pointwise errors for the registration of two of the patients in the dataset to the template and the average errors over all patients. Patient P\#091 serves as reference and it is not considered in the average errors calculation. In all cases, we compare the results obtained without and with a preliminary rigid registration of the geometries to the template by the Coherent Point Drift algorithm. The errors are quantified through the forward and backward local distances (FLD and BLD), expressed in \si{\centi\metre}. For reference, the template inlet diameter is $1.31$ \si{\centi\metre} for \emph{P\#091}.}
	
	\label{tab: RBF results}
	\def\arraystretch{1.5}
	\resizebox{.775\textwidth}{!}{
	\begin{tabular}{C{2cm} C{1.3cm} C{.1cm} C{1.35cm} C{1.35cm} C{.1cm} C{1.35cm} C{1.35cm}}
		\toprule
		& & &
		\multicolumn{2}{c}{\large\textbf{Max Errors}}&&
		\multicolumn{2}{c}{\large\textbf{Avg Errors}}\\
		
		\cmidrule(lr){4-5} \cmidrule(lr){7-8}
		
		& & & {\textbf{FCD}} & {\textbf{BCD}} && {\textbf{FCD}} & {\textbf{BCD}} \\
		
		\midrule 
		
		\multirow{3}{*}{\parbox{1.55cm}{\textbf{\centering Without\\Rigid\\Alignment}}} &
		
		\textbf{\emph{P\#090}} & & 0.2918 & 0.0777 && 0.2615 & 0.0770 \\
		&\textbf{\emph{P\#272}} & & 0.5760 & 0.0670 && 0.3428 & 0.0641 \\
		&\textbf{\emph{Average}} & & 0.7304 & 0.0861 && 0.4480 & 0.0759 \\
		
		\midrule
		
		\multirow{3}{*}{\parbox{1.55cm}{\textbf{\centering With\\Rigid\\Alignment}}} &
		
		\textbf{\emph{P\#090}} & & 0.2786 & 0.0736 && 0.2349 & 0.0739 \\
		&\textbf{\emph{P\#272}} & & 0.4667 & 0.0759 && 0.3002 & 0.0744 \\
		&\textbf{\emph{Average}} & & 0.6211 & 0.0825 && 0.3908 & 0.0756 \\
		
		\bottomrule
	\end{tabular}
}
\end{table}

Table~\ref{tab: RBF results} reports the results of the TPS interpolation algorithm, with and without prior rigid registration, obtained on two of the patients in the dataset (\emph{P\#090}, \emph{P\#272}) and averaged over all the shapes in the dataset (see Figure~\ref{fig: dataset} in the manuscript), except from \emph{P\#091}, that serves as reference. The pointwise registration errors, computed at all the cell centers of the available surface triangulations, are quantifies through the forward and backward local distances (FLD and BLD), expressed in \si{\centi\metre}. The former identifies the distance of each point in the mapped geometry from the closest one in the target, while the latter is the distance of each point in the target from the closest one in the mapped geometry. Figure~\ref{fig: RBF results} offers a visualization of the results, showing the positions of the interpolation points and the pointwise FLD values. A few considerations deserve attention. Firstly, TPS interpolation attains a notable degree of accuracy, with FLDs that are always well below the $1 \ \si{\centi\metre}$ threshold. Secondly, preliminary rigid registration is crucial when the original orientation of the target geometry differs from the reference one. This is showcased by patient \emph{P\#272}; indeed, the geometry obtained upon TPS interpolation without rigid registration is extremely irregular and convoluted, particularly in the aortic arch. Finally, we underline that TPS interpolation is rather sensitive to the values of ($i$) the smoothing parameter $w_H$ in Eq.\eqref{eq: TPS energy minimization}, and ($ii$) the tolerance $\tau$. The calibration of the latter is particularly important to obtain good quality results. In the reported tests, we select $\tau = 5 \cdot 10^{-3}$ for \emph{P\#090}, and $\tau = 2.5 \cdot 10^{-3}$ for \emph{P\#272}. However, to compute the aggregate metrics in Table~\ref{tab: RBF results}, we set $\tau = 5 \cdot 10^{-3}$ for all the geometries; this justifies why average errors are larger than patient--specific ones.

\subsection{Data augmentation pipeline}
\label{subsec: data augmentation pipeline}

\begin{algorithm}[t!]
	\caption{TPS--based data augmentation}
	\label{alg: data augmentation}
	\begin{algorithmic}[1]
		\Procedure{AugmentDataset}{N, $\mathcal{M}_1, \dots, \mathcal{M}_G$, $X_1, \dots, X_G$}       
		\Statex \Comment{\parbox[t]{.95\linewidth}{N: number of geometries; $\mathcal{M}_k$: k--th mesh; \\$X_k$: k--th set of sampling points}}
		
		\bigskip
		
		\State $n \gets 0$  
		\State $\mathcal{D} \gets [\mathcal{M}_1,\dots, \mathcal{M}_G]$  \Comment{Initialize the dataset}
		
		\medskip
		
		\While{$n < N$}      
		\State Sample $\alpha, \beta \sim \mathcal{U}(\{1,\dots,G\}), \ \alpha \neq \beta$   \Comment{Select two shapes}
		\State \emph{Ad hoc} rigid registration of $\mathcal{M}_\alpha$ to $\mathcal{M}_\beta$  
		\State CPD--based rigid registration of $\mathcal{M}_\alpha$ to $\mathcal{M}_\beta$  
		
		\medskip
		
		\State Sample $L \sim \mathcal{U}(\{1,2\})$   \Comment{Select number of vessel portions}
		\State Sample $L$ vessel portions $p_1, \dots, p_L$   \Comment{Select vessel portions}
		\State Sample $C_\ell \sim \mathcal{U}([0.5, 1])$, $\ell \in \{1, \dots, L\}$    \Comment{Select matching factors}
		
		\medskip
		
		\State $\tilde{X}_\alpha \gets \{{X_{\alpha, p_\ell}}\}_{\ell=1}^L$ \Comment{Interpolation points}
		\State $\tilde{X}_\beta \gets \{\{(1-C_\ell) X_{\alpha, p_\ell} +  C_\ell X_{\beta, p_\ell}\}\}_{\ell=1}^L$ \Comment{Interpolation field}
		\State $\mathcal{I} \gets \operatorname{TPS\_interpolator}(\tilde{X}_\alpha, \tilde{X}_\beta)$  
		\State $\mathcal{M}' \gets \mathcal{I}(\mathcal{M}_\alpha)$   \Comment{Query the interpolator}
		
		\medskip
		
		\If{$\operatorname{quality}(\mathcal{M}')$ is good}    \Comment{Check mesh quality}
		\State $\mathcal{D} \gets [\mathcal{D}, \mathcal{M}']$ \Comment{Update dataset}
		\State $n \gets n+1$
		\EndIf
		
		\medskip
		
		\EndWhile
		
		\medskip
		
		\Return $\mathcal{D}$
		
		\EndProcedure
	\end{algorithmic}
\end{algorithm}

The proposed TPS interpolation algorithm can provide good quality mapping results at a contained computational cost. However, robustness is a major drawback. Indeed, undesired artifacts are often introduced for too small values of $w_H$ (and for inadequate choices for $\tau$), while large values of $w_H$ negatively impact the overall goodness of fit. For this reason, we do not use TPS interpolation to solve the deformable registration problem on the vascular anatomies at hand, but we nonetheless exploit it for data augmentation. \\

Our TPS--based data augmentation pipeline is reported in Algorithm~\ref{alg: data augmentation}. The procedure involves the evaluation of ``partial'' TPS interpolators, where the word ``partial'' refers to the fact that only points from a subset of randomly selected vessel portions are considered. More specifically, at each iteration, we choose a random pair of geometries from the source cohort (line 5), whose corresponding sampled point sets are  $X_\alpha$, $X_\beta$, and a random number of vessel portions $L \in \{1,2\}$ (line 8). Firstly, we rigidly deform the points in $X_\alpha$ that belong to the selected vessel portions; this leads to the definition of $\bar{X}_\alpha = \bigcup_{\ell=1}^L \{\bar{X}_{\alpha, p_\ell}\}$ (lines 6,7). Then, the interpolation values are computed as follows (line 12):
\begin{equation*}
\tilde{X}_\beta := \bigcup_{\ell=1}^{L}\left\{(1-C_\ell) \bar{X}_{\alpha, p_\ell} +  C_\ell X_{\beta, p_\ell}\right\}~, \quad \text{with} \quad C_\ell \sim \mathcal{U}\left([0.5,1]\right)~.
\end{equation*}
Hence, the points selected from $X_{\alpha}$ are not mapped to the corresponding ones in $X_{\beta}$, but to some intermediate locations along the connecting segments, whose precise position depends on the random matching factors $C_\ell$. The derived TPS interpolator is used to deform the surface mesh $\mathcal{M}_\alpha$ of the first shape, so that a new triangulation $\mathcal{M}'$ is generated (lines 13,14). Finally, the resulting geometry is added to the dataset if the quality of the associated surface mesh is sufficiently high (line 15). Specifically, we require the scaled Jacobian --- the determinant of the Jacobian divided by the product of the two longest edges --- to be strictly positive for all cells and to have a bottom decile average value greater than $0.1$. From a qualitative point of view, this choice allows to obtain ``trustworthy'' geometries that do not feature undesired artifacts and irregularities. Incidentally, we remark that the obtained surface meshes are not used to perform numerical simulations, but only serve as a tool for shape discretization. Therefore, it is not necessary to require a high level of regularity, and we can accept the presence of a few bad elements.\\

Figure~\ref{fig: dataset}~(c) in the manuscript displays some of the shapes obtained by deforming the anatomies of four different patients with the proposed data augmentation pipeline. Despite being relatively simple, we remark the ability of the method to generate rather diverse shapes. Using Algorithm~\ref{alg: data augmentation}, we created $50$ new geometries from each of the ``original'' anatomies, hence assembling a dataset comprising $1,020$ shapes. However, the final dataset used to train and test the AD--SVFD model only counts $902$ geometries ($88.4\%$). The remaining $118$ ones have been manually removed, since they were showing artifacts that could not be captured with the implemented mesh quality check.  

\section{Hyperparameters tuning}
\label{app: HP tuning}
In this section, we focus on the calibration of the most relevant hyperparameters of the AD--SVFD model.

\subsection{ANN hyperparameters tuning}
\label{app: TPE tuning}
At first, we fine--tune the hyperparameters that are not related to the implicit neural representation of the source shapes. Since the latent codes are not involved in the calibration procedure, we can consider the case of a single shape--to--shape registration. This choice allows to dramatically lighten and speedup the training (from $\approx 8 \ \si{h}$ to $\approx 5 \ \si{min}$), hence enabling an exhaustive exploration of the hyperparameters' space at affordable computational costs. \\

We consider ten hyperparameters, namely: the activation function, width and depth of FA--NN ($\phi_{FA}$, $W_{FA}$, $L_{FA}$) and DF--NN ($\phi_{DF}$, $W_{DF}$, $L_{DF}$), the refinement level of the FPE ($N_e$), the penalty term $w_v$, the learning rate $\lambda = \lambda_\Theta$, and the adaptive sampling factor $a$. To limit the number of trainings and yet retain an extensive coverage of the hyperparameters' space, we run the Tree--structured Parzen Estimator (TPE) Bayesian algorithm~\cite{bergstra2011algorithms} for five different shapes (\emph{P\#090}, \emph{P\#144}, \emph{P\#188}, \emph{P\#207}, \emph{P\#272}) Considering quantized values, the total number of possible hyperparameters combinations is $8.64 M$. However, adopting TPE, we only perform $500$ trainings for each target shape; hence the overall duration of the fine--tuning procedure sets to $\approx 30 \ \si{h}$ per patient. \\

In order to identify a common (sub--)optimal set of hyperparameters, we marginalize the results of the five TPE runs with respect to the hyperparameter values. Firstly, for every patient, we associate every model with a score $s \in \R^+$, computed by averaging the mean forward and backward local distances associated with the direct and inverse mapping. To balance the contributions of the five patients, we normalize the model score $s$ by the best (i.e. the lowest) score $s^*$; this defines the normalized score $\tilde{s}$. Then, for each patient, every hyperparameter value is associated with the bottom decile average with respect to $\tilde{s}$, computed considering all the trained models that feature such value. Finally, for each hyperparameter value, we compute an aggregate performance score $S \in \R^+$ by averaging the bottom decile averages obtained on the five patients. Table~\ref{tab: TPE results} reports the results of the calibration procedure. \\

Even though the set of hyperparameters reported in Table~\ref{tab: TPE results} features (sub--)optimal properties for single shape--to--shape registration, those are not guaranteed to automatically transfer to the ``complete'' AD--SVFD model. In fact, with this configuration, the training of AD--SVFD fails, since all shape codes converge to the zero vector, leading to large errors. Empirically, we found that the problem is related to vanishing gradient issues in the trainable feature augmentation model compartment FA--NN. To circumvent this pitfall, we changed the FA--NN activation function $\phi_{FA}$ from \emph{ReLU} to \emph{leaky--ReLU} (with negative slope equal to $0.2$); this allowed to retain remarkable accuracy levels even in the multiple--shape scenario.

\begin{table}[!t]
	\caption{Results of the AD--SVFD hyperparameters calibration procedure. To save computational resources, we worked in a single shape--to--shape registration scenario and employed the Tree--structured Parzen Estimator algorithm, considering five different shapes. We refer to the text for a detailed definition of each hyperparameter. Every hyperparameter value is associated with the aggregate performance score $S$, computed from the average pointwise forward and backward local distances related to the direct and inverse mappings. Low values of $S$ correspond to accurate models. The optimal hyperparameter choices are marked in green. The yellow cells denote the hyperparameter values that were changed when incorporating the shape codes, for the simultaneous registration of multiple shapes. Notation: \emph{l--ReLU} stands for \emph{leaky--ReLU}, with a negative slope equal to $0.2$. }
	
	\label{tab: TPE results}
	\resizebox{\textwidth}{!}{
	\begin{tabular}{ C{2.5cm} | C{1.8cm} C{1.8cm} C{1.8cm} C{1.8cm} C{1.8cm} C{1.8cm}}
		\toprule
		
		\textbf{\normalsize Parameter} & & & & & & \Tstrut \\
		\midrule
		
		$\normalsize \bm{\phi}_{FA}$ & \cellcolor{green!25}$\bm{ReLU}$ & \cellcolor{yellow!25}$\bm{l-ReLU}$ & $\bm{ELU}$ & $\bm{SELU}$ & & \Tstrut \\
		& \cellcolor{green!25}1.0425 & \cellcolor{yellow!25}1.0655 & 1.0586 & 1.0573 & & \\
		\midrule
		
		$\normalsize \bm{\phi}_{DF}$ & $\bm{ReLU}$ & \cellcolor{green!25}$\bm{l-ReLU}$ & $\bm{ELU}$ & $\bm{SELU}$ & & \Tstrut  \\
		& 1.0632 & \cellcolor{green!25}1.0488 & 1.0578 & 1.0522 & & \\
		\midrule
		
		$\normalsize \bm{W_{FA}}$ & $\bm{2^3}$ & $\bm{2^4}$ & $\bm{2^5}$ & \cellcolor{green!25}$\bm{2^6}$ & $\bm{2^7}$ & \Tstrut  \\
		& 1.0502 & 1.0548 & 1.0531 & \cellcolor{green!25}1.0481 & 1.0758 & \\
		\midrule
		
		$\normalsize \bm{W_{DF}}$ & $\bm{2^6}$ & $\bm{2^7}$ & \cellcolor{green!25}$\bm{2^8}$ & $\bm{2^9}$ & & \Tstrut  \\
		& 1.0677 & 1.0489 & \cellcolor{green!25}1.0413 & 1.0715 & & \\
		\midrule
		
		$\normalsize \bm{L_{FA}}$ & $\bm{0}$ & $\bm{1}$ & $\bm{2}$ & \cellcolor{green!25}$\bm{3}$ & $\bm{4}$ & \Tstrut  \\
		& 1.0632 & 1.0627 & 1.0557 & \cellcolor{green!25}1.0474 & 1.0629 & \\
		\midrule
		
		$\normalsize \bm{L_{DF}}$ & $\bm{4}$ & \cellcolor{green!25}$\bm{5}$ & $\bm{6}$ & $\bm{7}$ & $\bm{8}$ & \Tstrut  \\
		& 1.0489 & \cellcolor{green!25}1.0444 & 1.0504 & 1.0709 & 1.0717 & \\
		\midrule
		
		$\normalsize \bm{N_e}$ & $\bm{0}$ & $\bm{1}$ & $\bm{2}$ & \cellcolor{green!25}$\bm{3}$ & $\bm{4}$ & $\bm{5}$ \Tstrut  \\
		& 1.0532 & 1.0639 & 1.0515 & \cellcolor{green!25}1.0336 & 1.0393 & 1.0604 \\
		\midrule
		$\normalsize \bm{w_v}$ & $\bm{10^{-6}}$ & \cellcolor{green!25}$\bm{10^{-5}}$ & $\bm{10^{-4}}$ & $\bm{10^{-3}}$ & $\bm{10^{-2}}$ & \Tstrut  \\
		& 1.0579 & \cellcolor{green!25}1.0371 & 1.0554 & 1.0654 & 1.0815 & \\
		\midrule
		
		$\normalsize \bm{\lambda_\Theta}$ & $\bm{10^{-4}}$ & $\bm{10^{-3.5}}$ & \cellcolor{green!25}$\bm{10^{-3}}$ & $\bm{10^{-2.5}}$ & $\bm{10^{-2}}$ & \Tstrut  \\
		& 1.1698 & 1.0671 & \cellcolor{green!25}1.0272 & 1.0565 & 1.1083 & \\
		\midrule
		
		$\normalsize \bm{a}$ & \bm{$0.00$} & \bm{$0.05$} & \bm{$0.10$} & \cellcolor{green!25}\bm{$0.15$} & \bm{$0.20$} & \bm{$0.25$} \Tstrut  \\
		& 1.0830 & 1.0535 & 1.0542 & \cellcolor{green!25}1.0404 & 1.0737 & 1.0738 \\
		
		\bottomrule
		
	\end{tabular}
}
\end{table}

\subsection{Shape code hyperparameters tuning}
\label{app: shape codes tuning}

\begin{table}[t!]
	\caption{Registration results of AD--SVFD considering different values of the regularization parameter $w_z$ and of the shape codes learning rate $\lambda_z$. In particular, we report the maximal pointwise errors on training and testing datapoints, obtained for six different values of $w_z$ and for five different values of $\lambda_z$. The errors are quantified through the forward and backward local distances (FLD and BLD), expressed in \si{\centi\metre}. The best value for each performance metric is marked in green. For reference, the template shape inlet diameter is $1.31$ \si{\centi\metre}, while the average inlet diameter in the dataset is $1.45$ \si{\centi\metre}.}
	
	\label{tab: results calibration z}
	\def\arraystretch{1.25}
	\resizebox{\textwidth}{!}{
	\begin{tabular}{C{1.8cm} C{1.35cm} C{1.35cm} C{1.35cm} C{1.35cm} C{1.35cm} C{1.35cm} C{1.35cm} C{1.35cm}}
		\toprule
		&
		\multicolumn{4}{c}{\large\textbf{Train errors (in cm)}}&
		\multicolumn{4}{c}{\large\textbf{Test errors (in cm)}}\\
		
		\cmidrule(lr){2-5} \cmidrule(lr){6-9}
		
		&
		\multicolumn{2}{c}{\large\textbf{Direct}}&
		\multicolumn{2}{c}{\large\textbf{Inverse}}&
		\multicolumn{2}{c}{\large\textbf{Direct}}&
		\multicolumn{2}{c}{\large\textbf{Inverse}}\\ 
		
		\midrule
		
		${\bm{w_z}}$
		& \textbf{FLD} & \textbf{BLD} & \textbf{FLD} & \textbf{BLD}
		& \textbf{FLD} & \textbf{BLD} & \textbf{FLD} & \textbf{BLD} \\
		
		\cmidrule(lr){2-3} \cmidrule(lr){4-5} \cmidrule(lr){6-7} \cmidrule(lr){8-9}
		
		$\bm{0.0}$ & \cellcolor{green!25}0.2095 & 0.2201 & \cellcolor{green!25}0.2583 & \cellcolor{green!25}0.2257 & 0.2725 & \cellcolor{green!25}0.1934 & \cellcolor{green!25}0.2422 & 0.2860 \\
		$\bm{10^{-5}}$ & 0.2333 & 0.2308 & 0.2834 & 0.2398 & \cellcolor{green!25}0.2714 & 0.2092 & 0.2564 & 0.2905 \\
		$\bm{10^{-4}}$ & 0.2166 & 0.2207 & 0.2719 & 0.2259 & 0.2853 & 0.2238 & 0.2815 & 0.3187 \\
		\cellcolor{green!25}$\bm{10^{-3}}$ & 0.2162 & \cellcolor{green!25}0.2175 & 0.2686 & 0.2297 & 0.2777 & 0.2253 & 0.2562 & \cellcolor{green!25}0.2642 \\
		$\bm{10^{-2}}$ & 0.2413 & 0.2353 & 0.3078 & 0.2408 & 0.3149 & 0.2751 & 0.3099 & 0.3713 \\
		$\bm{10^{-1}}$ & 0.3019 & 0.2794 & 0.3855 & 0.3020 & 0.4587 & 0.3641 & 0.4294 & 0.6546 \\
		
		\midrule
		
		${\bm{\lambda_z}}$
		& \textbf{FLD} & \textbf{BLD} & \textbf{FLD} & \textbf{BLD}
		& \textbf{FLD} & \textbf{BLD} & \textbf{FLD} & \textbf{BLD} \Tstrut \\
		
		\cmidrule(lr){2-3} \cmidrule(lr){4-5} \cmidrule(lr){6-7} \cmidrule(lr){8-9}
		
		$\bm{10^{-4}}$ & 0.2557 & 0.2404 & 0.3107 & 0.2680 & 0.4529 & 0.2679 & 0.3105 & 0.4771 \\
		$\bm{5 \cdot 10^{-4}}$ & \cellcolor{green!25}0.2072 & 0.2176 & \cellcolor{green!25}0.2665 & \cellcolor{green!25}0.2176 & 0.2783 & \cellcolor{green!25}0.2144 & 0.2671 & 0.3117 \\
		\cellcolor{green!25}$\bm{10^{-3}}$ & 0.2162 & \cellcolor{green!25}0.2175 & 0.2686 & 0.2297 & \cellcolor{green!25}0.2777 & 0.2253 & \cellcolor{green!25}0.2562 & \cellcolor{green!25}0.2642 \\
		$\bm{5 \cdot 10^{-3}}$ & 0.2574 & 0.2444 & 0.2938 & 0.2572 & 0.3556 & 0.2264 & 0.2893 & 0.3202 \\
		$\bm{10^{-2}}$ & 1.1374 & 2.1948 & 2.2775 & 1.1860 & 1.5228 & 1.8003 & 1.8297 & 1.5442 \\
		
		\bottomrule
	\end{tabular}
}
\end{table}

We focus on the calibration of two hyperparameters related to the shape codes, namely the regularization factor $w_z$ (see Eq.\eqref{eq: loss function}) and the learning rate $\lambda_z$. Table~\ref{tab: results calibration z} reports the maximal pointwise errors --- quantified through the forward and backward local distances FLD and BLD, in \si{\centi\metre} --- corresponding to different choices of $w_z$ (for $\lambda_z=10^{-3}$) and $\lambda_z$ (for $w_z=10^{-3}$). Concerning the regularization parameter, the results show little sensitivity, provided that sufficiently small values are considered. Indeed, all the models featuring $w_z \leq 10^{-3}$ yield similar results, but accuracy deteriorates for larger values. Conversely, the quality of the results heavily depends on the choice of the learning rate $\lambda_z$. Indeed, sensibly larger errors are obtained when either too small (e.g. $\lambda_z = 10^{-4}$) or too large (e.g. $\lambda_z = 10^{-2}$) values are selected. Ultimately, based on the obtained results, we set $w_z = 10^{-3}$ and $\lambda_z = 10^{-3}$. Notably, we choose $w_z$ as large as possible, in order to maximally regularize the latent space without compromising the registration accuracy.

\end{document}